\documentclass[11pt,final,onecolumn,journal]{IEEEtran}

\usepackage{ifpdf}
\usepackage{cite}

\usepackage{natbib}

\usepackage{booktabs} 

\usepackage{amsmath, amssymb}
\usepackage{array}
\usepackage{algorithmic}

\usepackage{bm}
\usepackage{color}

\usepackage{cleveref}
\usepackage{dsfont}
\usepackage[ruled,norelsize]{algorithm2e}

\usepackage{subfigure}
\usepackage{pgfplots} 
\pgfplotsset{compat=newest} 
\pgfplotsset{plot coordinates/math parser=false} 

\usepackage{url}

\newcommand{\E}{\mathsf{E}}
\newcommand{\Pagr}{\mathsf{P}^{\rm agr}}
\newcommand{\Pdis}{\mathsf{P}^{\rm dis}}
\newcommand{\Tr}{\operatorname{trace}}

\newcommand{\T}{\top}
\newcommand{\mc}{\mathcal}
\newcommand{\defeq}{\triangleq}

\newcommand{\1}{\mathds{1}}

\newcommand{\btheta}{\bm{\theta}}

\newcommand{\Ps}{\mc{P}}

\newcommand{\N}{\mc{N}}

\renewcommand{\Re}{\mathbb{R}}
\newcommand{\St}{\mathbb{S}}
\newcommand{\Ac}{\mathbb{A}}

\newcommand{\M}{\mathbb{M}}

\newcommand{\wrap}[1]{\begingroup\begin{tabular}{c} #1 \end{tabular}\endgroup}

\newtheorem{assumption}{Assumption}
\newtheorem{lemma}{Lemma}
\newtheorem{proposition}{Proposition}
\newtheorem{theorem}{Theorem}
\newtheorem{corollary}{Corollary}

\title{Fully Distributed Actor-Critic Architecture for Multitask Deep Reinforcement Learning}

\begin{document}

\author{
Sergio~Valcarcel~Macua, Ian Davies,
Aleksi Tukiainen, Enrique Munoz de Cote
\thanks{Work done while affiliated with Secondmind.}
\thanks{E-mail: {\tt \{sergiovalmac, davies.ian.r, aleksi.tukiainen\}@gmail.com, enrique@people-ai.com}.}
}

\maketitle

\begin{abstract}
We propose a fully distributed actor-critic architecture, named \textit{Diff-DAC}, with application to multitask reinforcement learning (MRL). During the learning process, agents communicate their value and policy parameters to their neighbours, diffusing the information across a network of agents with no need for a central station. Each agent can only access data from its local task, but aims to learn a common policy that performs well for the whole set of tasks. The architecture is scalable, since the computational and communication cost per agent depends on the number of neighbours rather than the overall number of agents. We derive Diff-DAC from duality theory and provide novel insights into the actor-critic framework, showing that it is actually an instance of the dual ascent method. We prove almost sure convergence of Diff-DAC to a common policy under general assumptions that hold even for deep-neural network approximations. For more restrictive assumptions, we also prove that this common policy is a stationary point of an approximation of the original problem. Numerical results on multitask extensions of common continuous control benchmarks demonstrate that Diff-DAC stabilises learning and has a regularising effect that induces higher performance and better generalisation properties than previous architectures.
\end{abstract}

%
\section{Introduction}
\label{sec:introduction}
%
%
%

\noindent Within a decade, billions of interconnected devices will be sensing, processing, and exchanging data throughout the global economy \citep{gartner2015}. 
Centralised reinforcement learning (RL) architectures, where multiple devices collect data that is sent to a central agent, may be infeasible at such a scale due to communication costs, 
excessive delays in communication, lack of resilience to link failures (e.g. consider a sensor network where the gateway that connects to the Internet fails), or even privacy concerns when sensitive data have to hop through multiple devices before reaching the central point of processing.

\textit{Diffusion}-based fully-distributed algorithms, offer a solution to these problems. Under such systems, agents learn from their local data and in the absence of a central coordinator. Agents can communicate only with their neighbours.
Therefore, the communication cost per agent scales linearly with its number of neighbours; 
all agents learn in parallel and are able to adapt to any number of neighbours, so the process is naturally more robust against device or link failure; 
and as the agents do not exchange data samples, privacy is naturally preserved. 

In addition, 
\textit{diffusion} schemes enjoy  theoretical convergence guarantees; and have shown the ability to outperform standard centralised schemes in the context of non-convex optimisation \citep[Ch. 4]{valcarcel2017phdthesis}.

The main motivation of this work is to show that the \textit{diffusion} mechanism can be used to develop fully distributed \textit{actor-critic} methods that enjoy the same favourable properties observed in distributed optimisation problems.
The proposed architecture, named \textit{diffusion-distributed-actor-critic} (Diff-DAC), is very flexible and in principle can be applied to any gradient-based single-agent method, as explained in Section \ref{sec:multiagent-approach}.

\subsection*{Related work}
The problem of learning policies that generalise across tasks is, in general, known as the multitask reinforcement learning (MRL) problem \citep{taylor2009transfer}.
There are multiple approaches to MRL. 
In the context of model-free deep-RL, one approach is to learn a single policy network that is able to control all tasks by learning shared low-level features \citep{parisotto2015actor}.
An alternative approach is to learn policies that are related but distinct for each task. This can be achieved architecturally, by learning modular policy networks \citep{andreas2017Modular}; by learning latent features that correlate the policies \citep{bou-ammar2014online,el2017scalable};
or by ensuring that the individual policies remain close to one another, e.g., their KL divergence or the Euclidean distance of their parameters are small \citep{teh2017distral}.
Model-based MRL approaches include learning a common model of the dynamics and then using this prior knowledge to solve new tasks \citep{fu2016one}.

Most previous MRL approaches are centralised, in the sense that they assume an agent with access to data from all observed tasks 
\citep{bou-ammar2014online,parisotto2015actor,teh2017distral,andreas2017Modular,fu2016one}.
But if the number of tasks is large or their data are \textit{geographically distributed},
the communication cost of transmitting the data to a central station, together with the sensitivity to link or node failure and other issues, may be prohibitive, as discussed above.

Previous works have scaled single-task reinforcement learning by distributing the sampling of training data. The Asynchronous Advantage Actor Critic (A3C) algorithm proposed by \cite{mnih2016asynchronous} enables reinforcement learning across multiple machines through the decentralised collection of data samples. Under A3C, decentralised agents pass locally calculated gradients to a central node that performs the global gradient update. Then, the decentralised agents periodically update their local policies by copying the parameters of the global policy maintained at the central node. \cite{espeholt2018impala} take an alternative approach by passing trajectories of experience in the place of calculated gradients. Both approaches have been shown to make training more stable and enable reinforcement learning agents to train faster and to higher final performance than traditional non-decentralised sampling schemes.

Our approach goes further and decentralises learning as well as sampling so that parameter sharing between agents is not centrally mediated. 
Furthermore, we develop our approach for the MRL problem and investigate the characteristics of policies learned in a decentralised manner.

The idea of scaling MRL with distributed optimisation was first proposed by \cite{el2017scalable} with the Distributed Multitask Policy Search (Dist-MTPS) method. 
Dist-MTPS assumes a factored policy model composed of two terms: a linear transformation that represents latent knowledge shared across all tasks, and a task-specific coefficient vector.
The dot product of these terms forms the parameter vector. This parameter vector linearly combines task-dependent features representing the state, to give the mean of a task-specific Gaussian distribution from which actions are drawn. The MRL problem is formulated as a non-convex optimisation problem over the linear transformation and the task-specific coefficients. A local optimum is then obtained with an iterative bi-level optimisation approach: at each iteration, the task-specific parameters are fixed and the agents find a consensus on the shared linear transformation, then the linear transformation is fixed and each agent finds its own task-specific parameter from local data. In order to find the shared linear transformation in a scalable manner, Dist-MTPS extends a distributed implementation of the alternating-direction-method-of-multipliers (ADMM) due to \cite{wei2012distributed}.

Our work improves over Dist-MTPS in a number of ways:
	\textit{i)} Dist-MTPS relies on linear function approximation, which requires finding salient features,
	and it only considers policies in the natural exponential family of distributions.
	Diff-DAC, on the other hand, allows nonlinear parametrisations, like deep learning architectures, which avoid costly feature engineering, and are able to learn more expressive policies.
	\textit{ii)} Distributed ADMM updates the agents in sequential order which requires finding a cyclic path that visits all agents.
	Finding such a path is generally an NP-hard problem \citep{karp1972reducibility}.
	Diff-DAC however, uses a diffusion strategy \citep{sayed2014adaptation}, 
	where each agent interacts with its neighbours, with no ordering, 
	and possibly asynchronously \citep{zhao2015asynchronousI}.
	\textit{iii)} Sequential updates, as in Dist-MTPS, are sensitive to agent or link failures, since they stop the information flow.
	Diffusion strategies, however, are robust to such failures as the agents can still operate even if parts of the network become isolated.

To the best of our knowledge, all other previous works on fully distributed RL consider only tabular or linear function approximations  \citep{Kar2012,valcarcel2013distributed,tutunov2016exact},
and do not apply immediately to expressive nonlinear approximations.
For example, \cite{Kar2012} added a consensus rule to tabular Q-learning. Its nonlinear extension therefore raises questions such as whether we should add a diffusion step to the target network updates. A principled response to this question would be an alternative contribution to our actor-critic approach. 
The Dist-GTD method presented by \cite{valcarcel2013distributed} is for policy evaluation with linear approximation, and its extension to control and nonlinear approximations is far from trivial even for the single-agent GTD.
Finally, \cite{tutunov2016exact} proposed a second order method, implying the inversion of the Hessian at every agent,
which might be problematic for neural networks with hundreds of neurons.
Other earlier  works suffer from similar drawbacks.

More recently\footnote{After the appearance of a preliminary version of this draft: \cite{valcarcel2017diffdac}.}, independent variants of the Diff-DAC architecture have been proposed for complementary scenarios. \cite{zhang2018fully} studied a similar algorithm adapted to the case where the agents interact with each other in the same environment. \cite{assran2019gossip} studied the computational efficiency, in terms of GPU utilisation, of a similar algorithm when all agents aim to solve the same task.

\subsection*{Contributions} 
\begin{enumerate}
\item We propose a fully distributed architecture, named Diff-DAC, that allows us to transform single-agent actor-critic algorithms into fully distributed implementations, which can be applied to the MRL problem, and scales gracefully to large number of tasks even with geographically distributed data.

\item We derive Diff-DAC from duality theory and provide novel insights into the standard actor-critic framework, showing that it is actually an instance of the dual ascent method.

\item We study the asymptotic convergence of the Diff-DAC architecture and prove that all agents converge to a common policy even for non-linear policy and value function representations. Under more restrictive assumptions, we also show that this common policy approximates a stationary point of the multitask RL objective.

\item We apply the Diff-DAC architecture to two algorithms: a simple Actor-Critic (SiAC) and the Advantage Actor Critic (A2C)---a synchronous version of A3C from \cite{mnih2016asynchronous}---, to study the stability and regularisation capabilities of the diffusion mechanism.

\item We perform multiple experiments over robotic control tasks, illustrating that even the Diff-DAC implementation of SiAC (Diff-SiAC) outperforms Dist-MTPS for fully distributed multitask RL;
that the Diff-DAC architecture is more stable and usually achieves better local optima than the centralised approach; and that Diff-DAC exhibits interesting generalisation properties.
\end{enumerate}

%
\section{Problem Formulation}
\label{sec:problem-formulation}
%

In this section, we formalise tasks as Markov decision processes (MDPs),
define a family of tasks and introduce the multitask optimisation problem. 

Consider a parametric family of MDPs defined over state-action sets, $\St$ and $\Ac$.
Each MDP of the family has a different state transition distribution, $\Ps_{\theta}: \St \times \St \times \Ac \mapsto \Re^+$, that depends on some parameter $\theta \in \Theta$,
where $\Theta$ is a measurable set, and $\Re^+$ denotes the non-negative real values.
We assume the reward function, $r: \St \times \Ac \times \St \mapsto \Re$,
and the distribution over the initial state, $\mu: \St \mapsto \Re^+$, are the same for all tasks.
A family of tasks is defined by a probability distribution over the parameter set, $f$, 
so that the parameter is a random variable\footnote{We use boldface font to denote random variables and regular font to denote instances or deterministic variables.}:
$\btheta = \theta \sim f$.
Let $\pi : \St \times \Ac \mapsto \Re^+$ be a stationary policy,
such that $\pi(a|s)$ denotes the conditional probability of taking action $a$ at state $s$.

For a given $\theta$, the single-task infinite-horizon discounted return objective is defined as:
\begin{IEEEeqnarray}{rCl}
	J_\theta(\pi)
&
\defeq  
&
	\E_{\Ps_\theta, \mu, \pi}
		\left[
			\sum_{t=0}^{\infty}
				\gamma^{t} 
				r ( \bm{s}_t, \bm{a}_t,  \bm{s}_{t+1})
		\right]
\notag\\
&=&
	\int_{\St}
		\rho_\gamma^\pi(s)
		\int_{\Ac}
			\pi(a|s)
			\int_{\St}
				\mc{P}_{\theta} (s'| s, a)
				r ( s, a, s')
				\ d s'
				\ d a
				\ d s
,
\label{eq:task-objective-function}
\end{IEEEeqnarray}
where 
$
	0 < \gamma <	1
$
is the discount factor,
and $\rho_\gamma^\pi$ is the discounted state measure\footnote{See \eqref{eq:discounted-state-measure-finite-state-action} below.} under policy $\pi$.

We consider the multitask RL problem of finding a single policy that performs well on average for the whole distribution of tasks. More formally, our goal is to solve the following problem:
\begin{IEEEeqnarray}{rCl}
\underset{  \pi }{\rm maximise} 	
\quad	
	J(\pi)
\defeq
	\int_\Theta
		f(\theta)
		J_\theta(\pi)
		\ d\theta
.
\label{eq:multitask-objective}
\end{IEEEeqnarray}

In practice, the agents will have to learn by only observing a subset of $N$ tasks that correspond to some parameters $\left \{ \theta_k \right\}_{k=1}^N$. 
Moreover, 
for large (or continuous) state-action sets, it is convenient to approximate the policy with a parametric function:
$\pi_w(a|s) \approx \pi(a|s)$, $\forall (s, a) \in \St \times \Ac$,
where $w \in \mathbb{W} \subseteq \Re^{M_\pi}$ is the parameter vector, the length of which is denoted by $M_\pi$.
Therefore, the actual problem the agents have to solve is to maximise the parametric empirical risk:
\begin{IEEEeqnarray}{rCl}
\underset{  w }{\rm maximise} 	
\quad	
	\overline{J}(w)
\defeq
	\frac{1}{N}
	\sum_{k=1}^N
		J_{\theta_k}(\pi_w)
.
\label{eq:empirical-risk}
\end{IEEEeqnarray}

Existence of a solution to \eqref{eq:empirical-risk} is guaranteed under standard assumptions on the policy set and state transition kernel, such that the induced state Markov chain has positive steady-state probabilities \citep{Bertsekas2012DPvol2}, as well as on the reward function such that the objective satisfies the Weirstrass conditions \citep[Prop. 3.2.1]{bertsekas2009convex}.

When all task parameters $\left \{ \theta_k \right\}_{k=1}^N$ are equal, \eqref{eq:empirical-risk} is the single-task RL problem;
when they differ, \eqref{eq:empirical-risk} becomes an MRL problem, where we aim to learn a single policy that performs well for the whole set of tasks. 
Although a single policy might perform well for some tasks but poorly for others,
experiments show that our solution to \eqref{eq:empirical-risk} can outperform Dist-MTPS, even when the latter learns related but distinct task-specialised policies.

%
\section{Networked Multi-Agent Setting and Diffusion Mechanism}
\label{sec:multiagent-approach}
%

In this section, we introduce the networked multi-agent setting wherein policies are learned in a fully distributed manner.

\begin{figure}
\begin{minipage}[c]{0.6\textwidth}
    \centering \includegraphics[width=0.96\columnwidth]{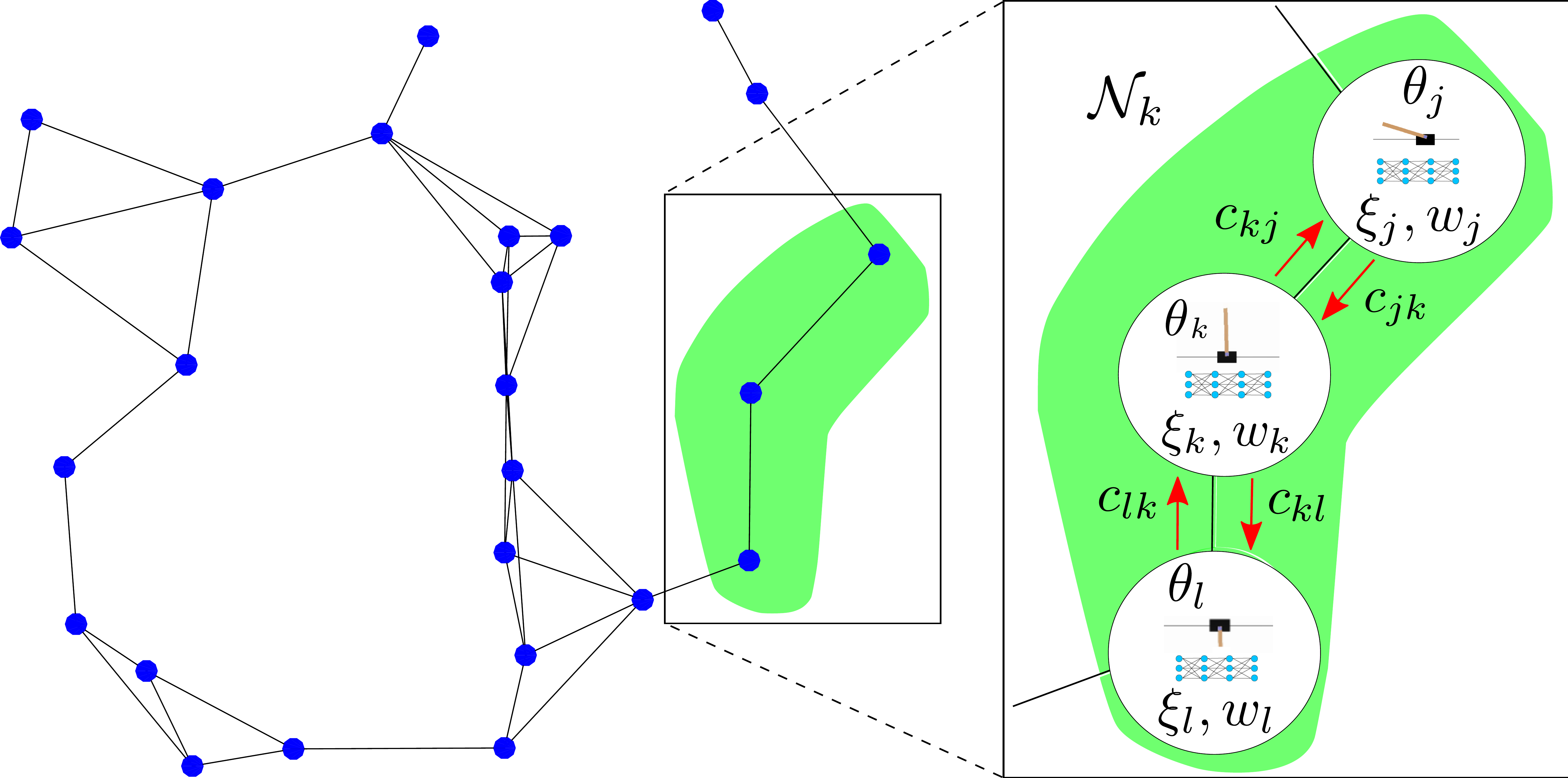}
\end{minipage}\hfill
\begin{minipage}[c]{0.38\textwidth}
    \caption{
       Example of network and detailed neighbourhood.
Blue nodes represent agents, and edges represent their connectivity.
On the right, the figure zooms in on neighbourhood $\N_k$ (the green area), 
where each agent $k$ runs its own instance of the environment (here illustrated as swing-up cart-poles, each with different pole length and mass).
As explained in Section \ref{sec:multiagent-approach}, agent $k$ transmits its critic and actor parameters, $\xi_{k,i}$ and $w_{k,i}$, to its neighbours $j$ and $l$;
and it receives their parameters $\xi_{j,i}, w_{j,i}$ and $\xi_{l,i}, w_{l,i}$, and combines them with weights $c_{jk}$ and $c_{lk}$, respectively.}
  \end{minipage} 
\end{figure}

We have a network of agents, each one learning from data coming from its own task.\footnote{For ease of exposition, 
we assume that each agent is allocated with one task, similar to \cite{el2017scalable}. 
The extension to multiple tasks per agent is trivial.}
Let $\N \defeq \{ 1,\ldots,N \}$ denote the set of agents.

The network is expressed as a graph, $\Gamma$, where nodes represent agents and edges represent communication links.
The graph can be represented by a non-negative matrix of size $N \times N$, denoted
%
$
	C 
\defeq
	\left( c_{lk} \right)_{l,k=1}^N
$,
such that the element $c_{lk} \ge 0$ represents the weight given by agent $k$ to information coming from $l$.
Each agent $k \in \N$ has data coming from its own task only, with parameter $\theta_k \sim f$;
and it is only allowed to communicate within its own neighbourhood, $\N_k$, 
which is defined as the set of agents to which it is directly connected, including $k$ itself:
\begin{IEEEeqnarray}{rCl}
	\N_k 
\defeq 
	\left\{
		l \in \{1,\ldots,N \}: c_{lk} > 0
	\right\}
.
\end{IEEEeqnarray}

We rely on the \textit{diffusion} mechanism for fully distributed optimisation \citep{Chen2013a_distributed, sayed2014adaptation}, which typically consists of two steps: local \textit{adaptation} and in-neighbourhood \textit{combination}.
During the adaptation step, each agent updates its parameters in the direction of its stochastic gradient calculated from local data.
In the combination step, each agent averages its local approximation with those coming from its neighbours.
For problem \eqref{eq:empirical-risk}, these two steps are described by the following updates, which run in parallel for all agents $k=1,\ldots,N$:
\begin{IEEEeqnarray}{rClC}
\IEEEnosubnumber
	\widehat{w}_{k,i+1}
&
=
&
	w_{k,i}
	+
	\beta_{i+1}
	\widehat{\nabla}_w J_{\theta_k}(\pi_{w_{k,i}})
\;\qquad
& {\text{(adaptation)}}
\IEEEyessubnumber \label{eq:adaptation}
\\
	w_{k,i+1}
&
=
&
	\sum_{l \in \N_k}
		c_{lk}
		\widehat{w}_{l,i+1}
,
& {\text{(combination)}}
\IEEEyessubnumber \label{eq:combination}
\end{IEEEeqnarray}
where $i$ is the iteration index;
$\beta_i$ is the step-size; 
$w_{k,i}$ is the approximate solution to the global problem \eqref{eq:empirical-risk} available at agent $k$ at iteration $i$;
$\widehat{\nabla}_w J_{\theta_k}(\pi_w)$ is the stochastic gradient of its local objective function evaluated at its current parameter $w_{k,i}$;
and $\widehat{w}_{k,i+1}$ is an intermediate parameter resulting from the local adaptation step.

In order to ensure that the information flows through the network, 
we assume that the graph, $\Gamma$, is \textit{strongly connected} (i.e., there is at least one path between every pair of agents),
and require the following standard conditions on the connectivity matrix $C$,
which together with strong connectivity make it doubly-stochastic and primitive
\citep{sayed2014adaptation,valcarcel2013distributed}:
\begin{IEEEeqnarray}{rCl}
	C^\T \1_N = \1_N
,\;
	C \1_N = \1_N
,\;\;
\text{ and }\;
	c_{lk} 
&
\ge
&	
	0
,\:\; \forall l,k \in \{ 1,\ldots,N \}
,\quad
\label{eq:non-negative-coefficients}
\\
	\Tr \:[C]
&
>
& 
	0
,\quad
\label{eq:aperiodic}
\end{IEEEeqnarray}
where $\1_N$ is a vector of ones of length $N$.
There are procedures for every agent $k$ to find the weights $\left\{ c_{lk} \right\}_{l\in\N_k}$ in a fully distributed manner,
such that the resulting $C$ satisfies \eqref{eq:non-negative-coefficients}--\eqref{eq:aperiodic}. 
One such procedure is the Hastings rule \cite[p.492]{zhao_performance_2012, sayed2014adaptation}.

%
\section{Architecture for Fully Distributed Actor-Critic Algorithms}
\label{sec:distributed-deep-actor-critic}
%

Equations \eqref{eq:adaptation}--\eqref{eq:combination} constitute the basis of a diffusion-based distributed policy gradient algorithm where we only optimise over the policy space, typically using an unbiased but high variance Monte Carlo estimate of the policy gradient, like REINFORCE \citep{williams1992simple}.

Actor-critic methods were introduced to reduce the variance of the policy gradient estimate  (see, e.g., \cite{GrondmanActorCritic2012} and references therein). 
In particular, it has been proven that an optimal control variate is given by the value function, which is defined for some given policy $\pi$ and some given task parameter $\theta$ as follows:
\begin{IEEEeqnarray}{rCl}
	v_\theta^\pi(s)
&
\defeq  
&
	\E_{\Ps_\theta, \pi}
		\left[
			\sum_{t=0}^{\infty}
				\gamma^{t} 
				r ( \bm{s}_t, \bm{a}_t,  \bm{s}_{t+1})
		\: 
		\big| 
		\:
		s_0 = s
		\right]
.
\label{eq:value-function}
\end{IEEEeqnarray}
In this context, the policy (used by the agent to operate in the environment) is known as the \textit{actor}, and the value function (which evaluates the goodness of a policy) is known as the \textit{critic}.

In this section, we propose a diffusion-based fully distributed actor-critic architecture, named Diff-DAC, where the agents cooperate to both estimate the multitask critic and optimise the actor.
We derive Diff-DAC from first principles, as a primal-dual scheme to find the saddle point of an approximate multitask Lagrangian. 
Our primal-dual derivation makes clear that the agents can benefit from applying diffusion to solve both the primal and dual problems in a cooperative manner. 
This is not obvious when the critic is motivated only from a variance reduction point of view.
Let's proceed to explain the details.

%
\subsection{Primal-dual derivation of actor-critic methods}
\label{ssec:primal-dual-derivation}
%
Throughout this subsection, we assume finite state-action sets for the sake of simplicity.
Hence, the single-task objective \eqref{eq:task-objective-function} can be expressed as a dot product:
\begin{IEEEeqnarray}{rCl}
    J_{\theta}(\pi) 
&=& 
    \mu^\T v^\pi_{\theta}
,
\label{eq:single-task-objective-as-dot-product}
\end{IEEEeqnarray}
where 
$
	v^{\pi}_{\theta}
\defeq  
	\left(
		v^{\pi}_{\theta}(s)
	\right)_{s \in \St}
\in 
\Re^{|\St|}
$,
is a vector with the value function for every state.

We can find the maximum of \eqref{eq:single-task-objective-as-dot-product} as the solution to the following linear program (LP) \citep[Sec. 9.1]{Puterman2005}:
\begin{IEEEeqnarray}{rCl}
\begin{aligned}
\underset{ v }{\rm minimise} 	
	&\;\;
		\mu^\T v
\\
{\rm s.t.}
	\quad\;\; &
		v(s)
	\ge 
		\sum_{s' \in \St}
		\Ps_{\theta} (s'|s,a) 
		\left(
			r(s,a,s')
			+
			\gamma
			v(s')
		\right)
	,\quad 
	    \forall (s,a) \in \St \times \Ac
.
\end{aligned}
\label{eq:linear-program}
\end{IEEEeqnarray}

The Lagrangian of \eqref{eq:linear-program} is given by:
\begin{IEEEeqnarray}{rCl}
	L_{\theta}(v,d)
&
=
&
	\mu^\T v
	+
	\sum_{(s,a)\in\St\times\Ac}
	d(s,a)
	\left(
		\sum_{s' \in \St}
        	\Ps_{\theta}(s'|s,a) 
        	\left(
		    	r(s,a,s')
				+
		        \gamma
        		v(s')
        	\right)
        	-
        	v(s)
	\right)
,
\label{eq:lagrangian}
\end{IEEEeqnarray}
where the dual variable,
$
	d 
\defeq
	\left(
		d(s,a)
	\right)_{(s,a)\in\St\times\Ac}
\ge 
	0
$,
is a non-negative vector of length $|\St||\Ac|$.
Let us introduce the multitask Lagrangian that integrates over the distribution of tasks:
\begin{IEEEeqnarray}{rCl}
	L(v,d)
&
\defeq
&
\int_\Theta
	L_{\theta}(v,d)
	f(\theta)
	\ d\theta
\notag\\
&=&
	\mu^\T v
	+
	\sum_{(s,a)\in\St\times\Ac}
	d(s,a)
	\left(
		\sum_{s' \in \St}
        \Ps(s'|s,a)     	
        \left(
	        r(s,a,s')
			+
    	    \gamma
			v(s')
		\right)
       	-
       	v(s)
	\right)
,
\label{eq:multitask-lagrangian}
\end{IEEEeqnarray}
where we have introduced a shortcut for the expected state transition distribution:
\begin{IEEEeqnarray}{rCl}
	\Ps(s'|s,a)
\defeq
	\int_\Theta
		\Ps_{\theta}(s'|s,a)
		f(\theta)
		\ d\theta
.
\end{IEEEeqnarray}
Note that \eqref{eq:multitask-lagrangian} can be thought of as the Lagrangian of another LP, similar to \eqref{eq:linear-program} but where  $\Ps_{\theta}$ has been replaced with $\Ps$:
\begin{IEEEeqnarray}{rCl}
\begin{aligned}
\underset{ v }{\rm minimise} 	
	&\;\;
		\mu^\T v
\\
{\rm s.t.}
	\quad\;\; &
		v(s)
	\ge 
		\sum_{s' \in \St}
		\Ps (s'|s,a) 
		\left(
			r(s,a,s')
			+
			\gamma
			v(s')
		\right)
	,\quad 
	    \forall (s,a) \in \St \times \Ac
.
\end{aligned}
\label{eq:multitask-linear-program}
\end{IEEEeqnarray}
Indeed, problem \eqref{eq:multitask-linear-program} corresponds to a single MDP with transitions given by $\Ps$, which represents a valid distribution, as shown by the following proposition.
%
%
\begin{proposition}
\label{proposition:cooperative-transition-matrix-stochastic}
$\Ps$ is a row-stochastic matrix.
\end{proposition}
%
%
%
%
\begin{IEEEproof}
Stochastic matrices lie in a compact convex set \citep[Th. 8.7]{horn1990matrix}.
Hence, their convex combination lies in the same set \citep[p.24]{boyd2004convex}.
\end{IEEEproof}
%

Since \eqref{eq:multitask-linear-program} satisfies Slater's condition, 
strong-duality holds \citep[Sec. 5.2.3]{boyd2004convex}, and optimal primal and dual variables are attained and form a saddle-point of \eqref{eq:multitask-lagrangian}:
\begin{IEEEeqnarray}{rCl}
	\min_{v} 
	\max_{d}
	L(v,d)
&
=
&
	L(v^{\star}, d^{\star})
=
	\max_{d}
	\min_{v} 
	L(v,d)
,
\label{eq:saddle-point}
\end{IEEEeqnarray}
where $v^{\star}$ and $d^{\star}$ denote optimal primal and dual variables\footnote{Note $v^{\star}$ is unique, while there might be multiple optimal dual variables.} of \eqref{eq:multitask-linear-program}, respectively.

There are multiple approaches to find a saddle-point that satisfies optimality condition \eqref{eq:saddle-point}.
We focus on the \textit{dual-ascent} scheme \citep{arrow1958studies}, which consists in alternating between:
\begin{enumerate}
    \item finding a primal solution, given the dual variable, and
    \item ascending in the direction of the gradient of the Lagrangian w.r.t. the dual variable, given the primal variable.
\end{enumerate}
%

First, we show how to update the \textit{primal} variable given $d$, that is:
\begin{IEEEeqnarray}{rCl}
	v
\leftarrow
	\arg\min_{v} L(v,d)
.
\end{IEEEeqnarray}
The Karush-Kuhn-Tucker (KKT) conditions are sufficient for optimality in convex problems that satisfy Slater's condition and have differentiable objective and constraints \citep[Sec. 5.5.3]{boyd2004convex}. 
These KKT conditions include 
the feasibility constraints, 
non-negativity of the dual variable for inequality constraints, 
\textit{complementary slackness},
and null gradient of the Lagrangian w.r.t. the primal variable.
Since problem \eqref{eq:multitask-linear-program} is linear, first-order conditions do not depend on $v$.
Thus, the only condition that depends on $v$ is \textit{complementary slackness}:
\begin{IEEEeqnarray}{rCl}
	d(s,a)
	\left(
	    \sum_{s' \in \St}
		\Ps(s'|s,a)
		\left(
			r(s,a,s')
			+
	    	\gamma
			v(s')
		\right)
		-
		v(s)
	\right)
=
	0
,
\quad
\forall (s,a)\in \St \times \Ac
.
\label{eq:complementary-slackness}
\end{IEEEeqnarray}
Similar to the standard single-task problem (recall Proposition \ref{proposition:cooperative-transition-matrix-stochastic}),
it can be shown \citep[Sec. 6.9]{Puterman2005} that our dual variable is the discounted state-action visitation measure:
\begin{IEEEeqnarray}{rCl}
	d(s,a)
&=&
	\sum_{j\in\St}\mu(j)
	\sum_{t=0}^\infty
	    \gamma^t
	     p(s_t = s, a_t = a| s_0 = j)
,
\end{IEEEeqnarray}
such that:
\begin{IEEEeqnarray}{rCl}
	\rho_\gamma^\pi(s) 
&=&
	\sum_{a\in\Ac} d(s,a)
\label{eq:discounted-state-measure-finite-state-action}
.
\end{IEEEeqnarray}
Therefore, finding $d$ allows us to obtain the corresponding policy:
\begin{IEEEeqnarray}{rCl}
	\pi(a|s)
=
	\frac{d(a,s)}{\sum_{a\in\Ac} d(s,a)}
.
\label{eq:policy-from-dual-variable}
\end{IEEEeqnarray}
Since 
$\pi(a|s) \ge 0$ 
and
$
    \sum_{s' \in \St}
    \Ps(s'|s,a)
    \left(
    	r(s,a,s')
    	+
    	\gamma
    	v(s')
    \right)
    -
    v(s)
\le 
    0
$
for all $(s, a) \in \St \times \Ac$, we conclude that the Bellman equation, typically used to derive the \textit{critic} in actor-critic methods, and given by:
\begin{IEEEeqnarray}{rCl}
	v(s)
&
=
&
	\sum_{a\in\Ac}
	\pi(a|s)
	\left(
		\sum_{s'\in\St}
		\Ps(s'|s,a)
		\left(
			r(s,a,s')
			+
			\gamma
			v(s')
		\right)
	\right)
,
\quad
	\forall s\in\St
,
\label{eq:multitask-bellman-equation}
\end{IEEEeqnarray}
is sufficient for \eqref{eq:complementary-slackness}, and thus to find the primal variable that optimises the Lagrangian.

Second, for the \textit{dual} variable, we simply perform gradient ascent in the Lagrangian, yielding a recursion of the form:
\begin{IEEEeqnarray}{rCl}
	d
&
\gets
&
	\left[
		d
		+
		\beta
		\nabla_d L(v,d)
	\right]^+
,
\label{eq:dual-gradient-ascent-update-of-cooperative-problem}
\end{IEEEeqnarray}
where $\beta$ is the step-size,
$[\cdot]^+$ denotes projection on the non-negative orthant,
and $\nabla_d $ denotes the gradient w.r.t. the dual variable $d$:
\begin{IEEEeqnarray}{rCl}
	\nabla_d L(v,d) 
&
= 
&
	\left(
		\frac{\partial L(v,d)}{\partial d(s,a)}
	\right)_{ (s,a) \in \St \times \Ac }
.
\label{eq:gradient-Lagrangian-wrt-dual-variable}
\end{IEEEeqnarray}
Interestingly, 
note that the partial derivatives of the Lagrangian in \eqref{eq:gradient-Lagrangian-wrt-dual-variable} are equal to the so-named \textit{advantage function}, $A:\St\times\Ac \rightarrow \Re$, which was originally motivated for learning to control continuous time systems approximated with small discrete time steps \citep{baird1993advantage} and, more recently, as a general variance reduction technique \citep{BhatnagarNaturalActorCritic2009}:
\begin{IEEEeqnarray}{rCl}
	\frac{\partial L(v,d)}{\partial d(s,a)}
=
	A(s,a)
&
\defeq
&
	\sum_{s' \in \St}
   	\Ps (s'|s,a) 
   	\left(
   		r(s,a,s')
		+
		\gamma
		v(s')
	\right)
	-
    v(s)
.
\label{eq:multitask-advantage}
\end{IEEEeqnarray}
Since by learning $d^{\star}$, we can obtain $\pi^{\star}$ from \eqref{eq:policy-from-dual-variable},
the recursion in \eqref{eq:dual-gradient-ascent-update-of-cooperative-problem} can be seen as an \textit{actor} update.

In summary, \eqref{eq:multitask-bellman-equation} and \eqref{eq:dual-gradient-ascent-update-of-cooperative-problem} define an \textit{actor-critic} method for the tabular setting, with no need for parametric policies.
This novel derivation shows that by finding a saddle-point of the multitask Lagrangian \eqref{eq:multitask-lagrangian}, 
we obtain a solution to the original problem \eqref{eq:multitask-objective}, as formally stated by the following proposition.
\begin{proposition}
Let $v^\star$ and $d^\star$ satisfy the saddle-point optimality condition \eqref{eq:saddle-point}. Then, they constitute a solution to the multitask problem \eqref{eq:multitask-objective}: 
\begin{enumerate}
\item [i)] 
$
	\mu^\T v^\star
=
	\max_\pi
		\int_\Theta
		f(\theta)
		J_\theta(\pi)
		\ d\theta
$.
\item [ii)] Let
$
	\pi^\star(a|s)
\defeq
	\frac{d^\star(a,s)}{\sum_{a\in\Ac}d^\star(s,a)}
$, 
	$\forall (s,a) \in \St \times \Ac$,
then 
$
	\pi^\star
	\in 
	\arg\max_\pi
	\int_\Theta
		f(\theta)
		J_\theta(\pi)
		\ d\theta
$.
\end{enumerate}
\end{proposition}
\begin{IEEEproof}
Consider $\Ps$ as the state transition of some MDP, then Theorem 6.9.4 from \cite{Puterman2005} states the following:
\begin{IEEEeqnarray}{rCl}
		\mu^\T
		v^\star
&
=
&
	\sum_{s,a \in \St \times \Ac}
		d^\star(s,a)
		\sum_{s'\in\St}
			\Ps(s'|s,a)
			r(s,a,s')
=
	\max_\pi
	\mu^\T
	v^\pi
,
\label{eq:optimal-primal-gives-optimal-policy}
\end{IEEEeqnarray}
where
$
	v^\pi
\defeq
	\E_{\Ps, \pi}
		\left[
			\sum_{t=0}^{\infty}
				\gamma^{t} 
				r ( \bm{s}_t, \bm{a}_t,  \bm{s}_{t+1})
		\right]
$.
From \eqref{eq:optimal-primal-gives-optimal-policy}, applying \eqref{eq:complementary-slackness} in \eqref{eq:multitask-lagrangian},
and using a relationship similar to \eqref{eq:single-task-objective-as-dot-product} but for the multitask transition kernel, $\Ps$, we have:
\begin{IEEEeqnarray*}{rCl+x*}
	L(v^\star, d^\star) 
&=&
	\max_\pi
		\mu^\T
		v^\pi
\\
&=&
	\max_\pi
		\int_{\St}
			\rho_\gamma^\pi(s)
			\int_{\Ac}
				\pi(a|s)
				\int_{\St}
					\Ps (s'| s, a)
					r ( s, a, s')
					\ d s'
					\ d a
					\ d s
\\
&=&
	\max_\pi
		\int_\Theta
		f(\theta)
		J_\theta(\pi)
		\ d\theta
.
& \nonumber\IEEEQEDhere
\end{IEEEeqnarray*}
\end{IEEEproof}

When we consider the empirical risk \eqref{eq:empirical-risk}, the expected state transition distribution, $\Ps$, can be approximated with this unbiased estimate: 
\begin{IEEEeqnarray}{rCl}
	\overline{\Ps}(s'|s,a)
&\defeq&
	\frac{1}{N} \sum_{k=1}^N 
		\Ps_{\theta_k}(s'|s,a)
,
\quad
	\forall (s, a, s') \in \left( \St \times \Ac \times \St \right)
. 
\end{IEEEeqnarray}
From Proposition \ref{proposition:cooperative-transition-matrix-stochastic}, we conclude that this is also a valid state transition distribution for some MDP.
Hence, the same arguments hold for this average distribution, and we can derive a dual-ascent method to obtain the saddle-point of an empirical approximation of the multitask Lagrangian \eqref{eq:multitask-lagrangian} that averages \eqref{eq:lagrangian} for the set of tasks 
$\left\{ \theta_k \right\}_{k=1}^N$:
\begin{IEEEeqnarray}{rCl}
	\overline{L}(v,d)
&=& 
	\frac{1}{N}
	\sum_{k=1}^N
		L_{\theta_k}(v,d)
.
\label{eq:empirical-lagrangian}
\end{IEEEeqnarray}
It is trivial to see that the only difference between \eqref{eq:multitask-lagrangian} and \eqref{eq:empirical-lagrangian} is that the former uses the expected distribution $\Ps$, while the latter uses its estimate $\overline{\Ps}$. 
This yields the following corollary.
\begin{corollary}
Let $v^\star$ and $d^\star$ be a saddle-point of \eqref{eq:empirical-lagrangian}. Then, they constitute a solution to the empirical risk problem: 
\begin{enumerate}
\item [i)] 
$
	\mu^\T v^\star
=
	\max_\pi
		\frac{1}{N}
		\sum_{k=1}^N
			J_{\theta_k}(\pi)
$.
\item [ii)] Let
$
	\pi^\star(a|s)
\defeq
	\frac{d^\star(a,s)}{\sum_{a\in\Ac}d^\star(s,a)}
$, 
	$\forall (s,a) \in \St \times \Ac$,
then 
$
	\pi^\star
	\in 
	\arg\max_\pi
		\frac{1}{N}
		\sum_{k=1}^N
			J_{\theta_k}(\pi)
$.
\end{enumerate}
\end{corollary}

In the following subsections, we propose a diffusion-based fully distributed actor-critic architecture for large (possibly infinite) state-action sets, with parametric value function and policy that approximate the saddle point of \eqref{eq:empirical-lagrangian}.

%
\subsection{Distributed critic}
%
%

In the previous subsection, we have seen that the primal update of the dual-ascent method
is equivalent to finding a value function that satisfies the multitask Bellman equation \eqref{eq:multitask-bellman-equation}.

When computing this value function for large or continuous state sets, it is common\footnote{See, e.g., \cite{ng1999policy,KondaActorCritic2003,MeloFitted2008, BhatnagarNaturalActorCritic2009, Powell2011, van2012reinforcement, weinstein2012bandit, wierstra2014natural, Lillicrap2015Continuous, heess2015learning, schulman2015high}.} to rely on some parametric function approximation: $v_{\xi}(s) \approx v(s)$, 
where $\xi \in \Re^{M_v}$ is the parameter vector of length $M_v$.
Thus, for some given policy, $\pi$, we can learn value function parameters $\xi$ by transforming \eqref{eq:multitask-bellman-equation} into a regression problem:
\begin{IEEEeqnarray}{rCl}
\underset{\xi}{\rm minimise} 	
&&
\quad
	\overline{J}(\xi)
\defeq
		\E_{\rho_\gamma^\pi}
	    \left[
	      \left (
	          v_{\xi} \left( \bm{s}_t \right)
	          -
	          \overline{\bm{y}}_t
	      \right )^2
	    \right]
,
\label{eq:distributed-expected-bellman-error}
\end{IEEEeqnarray}
where the target values are given by:
\begin{IEEEeqnarray}{rCl}
	\overline{\bm{y}}_t
&
\defeq
&
	\int_{\Ac}
		\pi(a|s)
		\int_{\St}
	    	\overline{\Ps} (s'|\bm{s}_t, a_t) 
	    	\left(
	   			r(\bm{s}_t, a_t, s') 
				+
		    	\gamma
				v_{\xi}(s')
			\right)
		\ d s'
	\ d a
,
\end{IEEEeqnarray}
This is equivalent to finding $\xi$, such that the advantage function is as close to zero as possible.

In order to derive a diffusion-based distributed critic, 
we have to reformulate the problem as minimising the average cost over all tasks.
The cost for each individual task takes the form:
\begin{IEEEeqnarray}{rCl}
	J_k(\xi)
&
\defeq
&
	\E_{\rho_\gamma^\pi}
	\left[
		\left (
			v_{\xi} \left( \bm{s}_{t} \right)
			-
			\bm{y}_{k,t}
      \right )^2
	\right]
,\;
	k=1,\ldots,N
,
\label{eq:individual-critic-cost}
\end{IEEEeqnarray}
where $\bm{y}_{k,t}$ is the target for the $k$-th task at time $t$:
\begin{IEEEeqnarray}{rCl}
	\bm{y}_{k,t}
&
\defeq
&
	\int_{\Ac}
		\pi(a|s)
		\int_{\St}
		   	\Ps_{\theta_k} (s'|\bm{s}_{t}, a_{t}) 
		   	\left(
				r(\bm{s}_{t}, a_{t}, s')
				+ 
			    \gamma 
				v_{\xi}(s')
		   	\right)
	   	\ d s'
   	\ d a
,
\end{IEEEeqnarray}
such that:
$
	 \overline{\bm{y}}_t
=
	1/N
	\sum_{k=1}^N
	\bm{y}_{k,t}
$.
%

We can use  Jensen's inequality to upper bound $J(\xi)$ by another function, 
$
	\widetilde{J}(\xi)
$, and use this upper bound as surrogate cost:
\begin{IEEEeqnarray}{rCl}
	\widetilde{J}(\xi)
&
\defeq
&
	\frac{1}{N}
	\sum_{k=1}^N
	J_k(\xi)
=
	\frac{1}{N}
	\sum_{k=1}^N
	\mathbb{E}
    \left[
		\left (
			v_{\xi} \left( \bm{s}_{t} \right)
			-
			\bm{y}_{k,t}
      \right )^2
    \right]
\ge 
	\mathbb{E}
    \left[
		\left (
			\frac{1}{N}
			\sum_{k=1}^N
				\left(
					v_{\xi} \left( \bm{s}_{t} \right)
					-
					\bm{y}_{k,t}
				\right)
      \right )^2
    \right]
=
	\overline{J}(\xi)
.
\qquad
\end{IEEEeqnarray}

Now, we can apply \textit{diffusion} to minimise $\widetilde{J}(\xi)$ in a distributed fashion, with every agent $k=1,\ldots,N$, applying \textit{adaptation} and \textit{combination} steps in parallel:
\begin{IEEEeqnarray}{rCl}
\IEEEnosubnumber
	\widehat{\xi}_{k,i+1}
&
=
&
	\xi_{k,i}
	-
	\alpha_{i+1}
	\widehat{\nabla}_\xi J_k (\xi_{k,i})
\IEEEyessubnumber \label{eq:value-adaptation}
\\
	\xi_{k,i+1}
&
=
&
	\sum_{l \in \N_k}
		c_{lk}
		\widehat{\xi}_{l,i+1}
,
\IEEEyessubnumber \label{eq:value-combination}
\end{IEEEeqnarray}
where 
$i$ is the iteration index;
$\alpha_i$ is the step-size; 
$\widehat{\nabla}_\xi J_k (\xi_{k,i})$ is the stochastic gradient evaluated at $\xi_{k,i}$
and estimated from a batch of $T_{k,i}$ local samples, $\left\{ \left( s_{k,t}, a_{k,t}, r_{k,t+1}, s_{k,t+1} \right) \right \}_{t=0}^{T_{k,i}}$, drawn from the stationary distribution induced by policy $\pi$.

Equations \eqref{eq:value-adaptation}--\eqref{eq:value-combination} represent a template for a fully distributed critic update.
Depending on how the targets are estimated, the sampling process and other factors, we can build different distributed critic methods. 
In addition, instead of using a simple stochastic gradient update in the adaptation step \eqref{eq:value-adaptation}, we can use more sophisticated methods, like momentum, adaptive learning rate, and other variants.
In this paper, we will introduce two specific algorithms that use this template, Diff-SiAC and Diff-A2C. But before discussing their details, we present a template for the fully distributed actor update.

%
\subsection{Distributed actor}
%
%
Recall from Sec. \ref{ssec:primal-dual-derivation} that the dual step of the dual-ascent method involves ascending in the direction of the gradient of the Lagrangian w.r.t. the dual variable.

For large or continuous state-action sets, it is convenient to approximate the policy with a parametric function.
Let $\pi_w \approx \pi$ denote the parametric approximation of the actual policy,
where $w \in \Re^{M_\pi}$ is the parameter vector of length $M_\pi$.

Replacing $\pi$ with $\pi_w$ in \eqref{eq:empirical-lagrangian},
we obtain an approximate Lagrangian,
$
	\widetilde{L}(v,w)
\approx
	\overline{L}(v,d)
$:
\begin{IEEEeqnarray}{rCl}
	\widetilde{L}(v,w)
&
=
&
	\mu^\T v
	+
	\int_\St \int_\Ac
	d_w(s,a)
	\overline{A}(s,a)
	\ ds\ da
,
\label{eq:approximate-lagrangian}
\end{IEEEeqnarray}
where $d_w(s,a)$ is the resulting approximation of the dual variable:
\begin{IEEEeqnarray}{rCl}
	d_w(s,a)
&
\defeq
&
	\pi_w(a|s)
	\rho_\gamma^{\pi_w}(s)
,
\end{IEEEeqnarray}
and 
$
\overline{A}(\cdot)
\defeq
	1/N
	\sum_{k=1}^N
		A_k(\cdot)
$
refers to the average advantage function, with: 
\begin{IEEEeqnarray}{rCl}
    A_k(s,a)
&
\defeq
&
	\int_{\St}
       	\Ps_{\theta_k} (s'|s,a) 
       	\left(
			r(s,a,s')
			+
        	\gamma
        	v(s')
        \right)
	d s'
       	-
       	v(s)
.
\label{eq:individual-advantage}
\end{IEEEeqnarray}
Thus, in order to approximate the optimal dual variable,
we can move in the ascent direction of the gradient of \eqref{eq:approximate-lagrangian} w.r.t. the policy parameter.
The following theorem provides the required gradient.
%
%
\begin{theorem}
The gradient of the Lagrangian w.r.t. the policy parameter is given by:
\begin{IEEEeqnarray}{rCl}
	\nabla_w \widetilde{L}	(v,w)
=
	\int_{\St} 
		\rho_\gamma^{\pi_w}(s)
		\int_{\Ac}
			\pi_{w}(a|s)
			\nabla_{w} \log \pi_{w}(a|s)
			\overline{A}(s,a)
	\ ds\ da
.
\label{eq:gradient-approximate-lagrangian}
\end{IEEEeqnarray}
\label{theorem:policy-gradient}
\end{theorem}
%
%
\begin{IEEEproof}
See Appendix \ref{app:proofs}.1.
\end{IEEEproof}

Interestingly, \eqref{eq:gradient-approximate-lagrangian} is similar to previous \textit{policy gradient} theorems \citep{Sutton1999policygradient}, 
with the important difference that it yields the advantage function explicitly;
while previous works motivated the advantage---as opposed to the state-action value function---as a variance reduction technique
\citep{williams1992simple,BhatnagarNaturalActorCritic2009}.

In order to derive a fully \textit{distributed} actor, we introduce the approximate Lagrangian for each individual task:
\begin{IEEEeqnarray}{rCl}
	\widetilde{L}_k(v,w)
&
\defeq
&
	\mu^\T v
	+
	\int_\St \int_\Ac
		\rho_\gamma^{\pi_w}(s)
		\pi_w(a|s)
		A_k(s,a)
	\ ds\ da
,
\label{eq:individual-approximate-lagrangian}
\end{IEEEeqnarray}
such that 
$
	\widetilde{L}(\cdot)
=
	1/N
	\sum_{k=1}^N
		\widetilde{L}_k(\cdot)
$.

Similar to the critic, we can apply diffusion to perform the actor update,
with step-size $\beta_{i+1}$:
\begin{IEEEeqnarray}{rCl}
\IEEEnosubnumber
	\widehat{w}_{k,i+1}
&
=
&
	w_{k,i}
	+
	\beta_{i+1}
	\widehat{\nabla}_w \widetilde{L}_k(v_{\xi_{k,i}},w_{k,i})
,
\IEEEyessubnumber \label{eq:policy-adaptation}
\\
	w_{k,i+1}
&
=
&
	\sum_{l \in \N_k}
		c_{lk}
		\widehat{w}_{l,i+1}
,
\IEEEyessubnumber \label{eq:policy-combination}
\end{IEEEeqnarray}
where $\widehat{\nabla}_w \widetilde{L}_k(v_{\xi_{k,i}},w_{k,i})$ is an estimate of each agent's local policy gradient---which is similar to 
\eqref{eq:gradient-approximate-lagrangian} but replacing the average advantage function, $\overline{A}$, with each agent's estimate of its local advantage function, $A_k$---evaluated at its local critic estimate, $v_{\xi_{k,i}}$, obtained from \eqref{eq:value-adaptation}--\eqref{eq:value-combination}.

Similar to the critic update, there are many ways of estimating the stochastic gradient in \eqref{eq:policy-adaptation}. 
Depending on the way of estimating the advantage function (see \cite{schulman2015high} for multiple estimators); the sampling process; and how the value parameters, $\xi_{k,i}$, relate to the policy parameters, we can obtain different fully distributed actor updates. 
These options lead to different actor-critic methods.

In the following subsections, we apply the Diff-DAC architecture to propose two actor-critic algorithms, which will allow us to demonstrate the benefits of Diff-DAC in terms of stabilisation, performance, generalisation and robustness.

\subsection{Simple Actor Critic (SiAC)}
\label{ssec:vac}

The first algorithms we introduce are centralised and Diff-DAC implementations of a simple actor-critic (SiAC) algorithm, characterised by having a very simple and highly biased estimator of the advantage function that allows us to evaluate the stability provided by the diffusion mechanism.
In particular, the parameters are only updated once the agent has completed an episode on each task.
In order to be sample efficient, we perform as many updates to both critic and actor as available transitions in the episode. That is, if at the $i$-\textit{th} iteration, the $k$-\textit{th} agent samples an episode with $T_{k,i}$ transitions, then we obtain a set of $T_{k,i}$ advantage estimates given by:
\begin{IEEEeqnarray}{rCl}
    \widehat{A}_{k,t}
&
\defeq
&
	\sum_{j=t}^{T_{k,i}}
		\gamma^{j-t}
		r_{k,j+1}
		-
		v_{\xi_{k,i}}(s_{k,t})
,\quad
0 \le t < T_{k,i}-1.
\label{eq:advantage-estimator}
\end{IEEEeqnarray}

Note that the advantage estimates $\left\{ \widehat{A}_{k,t} \right\}_{t=1}^{T_{k,i}}$ will be highly correlated, and can therefore cause learning to become unstable. 
This is done by design in order to conveniently study the stabilisation properties of the Diff-DAC architecture.

In order to compare the centralised and Diff-DAC architectures, we build two algorithms \textit{Centralised SiAC} and \textit{Diff-SiAC} where we introduce the following two changes: 
\textit{i)} Centralised SiAC has a single set of value and policy parameters whereas Diff-SiAC has $N$ sets of value and policy parameters (one per agent); and 
\textit{ii)} Centralised SiAC updates its set of value and policy parameters with advantage estimates calculated across all tasks whereas Diff-SiAC updates each agent's parameters with its own local set of samples and then combines parameters with diffusion.

For Centralised SiAC, the value and policy parameters, $\xi_i$ and $w_i$, are updated with stochastic gradients that average all samples and all advantage estimates from all tasks:
\begin{IEEEeqnarray}{rCl}
    \widehat{\nabla}_\xi \widetilde{J} (\xi_i, \pi_{w_i})
&=&
	\frac{1}{T_i}
	\sum_{k=1}^N
		\sum_{t=0}^{T_{k,i} - 1}
            \nabla_\xi v_{\xi_i}(s_{k,t})
            \widehat{A}_{k,t}
,
\label{eq:sgd-centralised-value}
\\
	\widehat{\nabla}_w \widetilde{L} (v_{\xi_{i}},w_{i})
&=&
	\frac{1}{T_i}
	\sum_{k=1}^N
		\sum_{t=0}^{T_{k,i} - 1}
		\nabla_{w} \log \pi_{w_{i}}(a_{k,t}|s_{k,t})
 		\widehat{A}_{k,t}
,
\label{eq:sgd-centralised-policy}
\end{IEEEeqnarray}
where $T_i \defeq \sum_{k=1}^N T_{k,i}$, denotes the total number of samples from the $i$-th episode of all tasks.

For Diff-SiAC, each agent updates its parameters, $\xi_{k,i}$ and $w_{k, i}$, with stochastic gradients that average its local samples and advantage estimates:
\begin{IEEEeqnarray}{rCl}
    \widehat{\nabla}_\xi J_k (\xi_{k,i}, \pi_{w_{k, i}})
&=&
	\frac{1}{T_{k,i}}
		\sum_{t=0}^{T_{k,i} - 1}
            \nabla_\xi v_{\xi_{k,i}}(s_{k,t})
            \widehat{A}_{k,t}
,
\label{eq:sgd-diff-dac-value}
\\
	\widehat{\nabla}_w \widetilde{L}_k (v_{\xi_{k, i}},w_{k, i})
&=&
	\frac{1}{T_{k,i}}
	\sum_{t=0}^{T_{k,i} - 1}
		\nabla_{w} \log \pi_{w_{k,i}}(a_{k,t}|s_{k,t})
 		\widehat{A}_{k,t}
.
\label{eq:sgd-diff-dac-policy}
\end{IEEEeqnarray}

Whilst we have considered a standard stochastic gradient for the derivations, in practice we are free to use any gradient-based optimiser to perform parameter updates. 
In the experiments with Centralised SiAC and Diff-SiAC, we use the Adam optimiser \citep{kingma2015adam}.

\subsection{Diffusion A2C (Diff-A2C)}

In this subsection, we introduce a Diff-DAC implementation of A2C that will allow us to evaluate a new regularisation effect that emerges from fully distributed sampling and learning.

A2C is a variant of the A3C algorithm described in Sec. \ref{sec:introduction}
that uses distributed sampling but synchronous learning updates.
In addition, A2C typically uses a more advanced multi-step advantage estimator that reduces variance and can lead to faster learning:
\begin{IEEEeqnarray}{rCl}
	\widehat{A}_{k,t}
& 
\defeq 
&
	\sum_{j=t}^{T_{k, i}}
		\gamma^{j-t}
		r_{k,j+1}
		+\gamma^{T_{k,i}}
		v_{\xi_{k,i}}(s_{k,T_{k, i}})
		- v_{\xi_{k,i}}(s_k, j)
,\quad
0 \le t < T_{k}-1,
\label{eq:a2c-advantage}
\end{IEEEeqnarray}
where $T_{k,i}$ is a parameter and denotes the number of transitions per parameter update. 
Note that the multi-step estimator \eqref{eq:a2c-advantage} allows A2C agents to set $T_{k,i}$ to a fixed small value such that the parameters are updated at a fixed interval in terms of environment steps, instead of at the end of each episode. 
The distributed sampling approach, the multi-step advantage estimates, and the quick updates allows A2C to offer stable and sample efficient learning.
In typical A2C approaches, the training of a single set of parameters is centralised. 

Since the advantage estimator in \eqref{eq:a2c-advantage} relies on local samples and parameters, we are able to apply it to agents in the Diff-DAC architecture, with stochastic gradients similar to \eqref{eq:sgd-centralised-value}--\eqref{eq:sgd-diff-dac-policy}. 
This yields an algorithm we refer to as Diff-A2C wherein both experience collection and training are decentralised. 

As with SiAC and Diff-SiAC we are free to choose the gradient-based optimiser. For the experiments with A2C and Diff-A2C we use RMSProp \citep{tieleman2012lecture} to demonstrate Diff-DAC working with varied optimisers.\footnote{During experimentation we observed similar results across runs with RMSProp and Adam optimisers.}
See Appendix \ref{app:pseudocode} for pseudocode and a discussion contrasting Diff-SiAC and Diff-A2C.

%
\section{Convergence Analysis}
\label{sec:convergence-analysis}
%

In this section, we illustrate some asymptotic properties of the distributed actor-critic iterations, \eqref{eq:value-adaptation}--\eqref{eq:value-combination} and \eqref{eq:policy-adaptation}--\eqref{eq:policy-combination}.
We provide two asymptotic properties of the Diff-DAC architecture. 
First, we show that all agents converge to an average recursion. 
Second, we show that this recursion converges to a stationary point of the parametric approximation of the MRL problem.

Let us start the analysis by rewriting the updates as follows:
\begin{IEEEeqnarray}{rCl}
\IEEEnosubnumber
	\widehat{\bm{\phi}}_{k,i+1}
&
=
&
	\bm{\phi}_{k,i}
	-
	\alpha_{i+1}
	\left(
		g_k\left(\bm{\phi}_{k,i}\right)
		+
		\bm{F}_{k,i+1}
	\right)
\IEEEyessubnumber 
\label{eq:joint-adaptation}
\\
	\bm{\phi}_{k,i+1}
&
=
&
	\sum_{l \in \N_k}
		c_{lk}
		\widehat{\bm{\phi}}_{l,i+1}
,
\IEEEyessubnumber 
\label{eq:joint-combination}
\end{IEEEeqnarray}
where we aggregated the critic and actor parameters into a single vector, and we expressed the stochastic gradient as the sum of the exact expected gradient, $g_k$, plus a random vector, $\bm{F}_{k,i+1}$:
\begin{IEEEeqnarray}{C}
	\bm{\phi}_{k,i}
\defeq
\left(
	\begin{array}{c}
	\bm{\xi}_{k,i} \\
	\bm{w}_{k,i}
	\end{array}
\right)
,\quad
	g_k\left(\bm{\phi}_{k,i}\right)
\defeq
\left(
	\begin{array}{c}
		\nabla_\xi J_k (\bm{\xi}_{k,i}, \pi_{\bm{w}_{k,i}})
	\\
		\frac{\beta_{i+1}}{\alpha_{i+1}}
		\nabla_w \widetilde{L}_k(v_{\bm{\xi}_{k,i}},\bm{w}_{k,i})
	\end{array}
\right)
,
\\
	\bm{F}_{k,i+1}
\defeq
\left(
	\begin{array}{c}
		\widehat{\nabla}_\xi J_k (\bm{\xi}_{k,i},\pi_{\bm{w}_{k,i}})
	\\
		\frac{\beta_{i+1}}{\alpha_{i+1}}
		\widehat{\nabla}_w \widetilde{L}_k(v_{\bm{\xi}_{k,i}},\bm{w}_{k,i})
	\end{array}
\right)
	-
	g_k\left(\bm{\phi}_{k,i}\right)
.
\end{IEEEeqnarray}

We require the following standard assumptions:
\begin{assumption}
\label{ass:connectivity-matrix}
The graph $\Gamma$ is strongly connected and $C$ satisfies conditions 
\eqref{eq:non-negative-coefficients}--\eqref{eq:aperiodic}.
\end{assumption}
\begin{assumption}
\label{ass:lipschitz-gradients}
Gradients $g_k$ are Lipschitz for all $k\in\N$:
\begin{IEEEeqnarray}{rCl}
	\left\|
		g_k (\phi)
		-
		g_k (\phi')
	\right\|
&
\le
&
	L_1 \left \| \phi - \phi' \right \|
,\quad
	0 < L_1 < \infty
,\;\;
	\forall \phi, \phi' \in \Re^{M_v+M_\pi}
.
\end{IEEEeqnarray}
\end{assumption}
\begin{assumption}
\label{ass:stepsizes}
Step size sequences $\left\{ \alpha_i \right\}_{i=0}^\infty$ and $\left\{ \beta_i \right\}_{i=0}^\infty$ consist of positive scalars satisfying:
\begin{IEEEeqnarray}{rCl}
	\sum_{i=0}^\infty
		\alpha_i
	=
	\sum_{i=0}^\infty
		\beta_i
,\quad
	\sum_{i=0}^\infty
		\left(
			\alpha_i^2
			+
			\beta_i^2
		\right)
<
	\infty
,\quad
    \lim_{i\rightarrow\infty}
	    \frac{\beta_i}{\alpha_i}
=
	0
.
\label{eq:stepsizes}
\end{IEEEeqnarray}
\end{assumption}
\begin{assumption}
\label{ass:square_integrable}
$
\left\{
	\bm{F}_{k,i+1}
\right\}_{i=0}^\infty
$ are square-integrable w.r.t. the increasing families of $\sigma$-fields
$
	\mc{F}_{k,i}
\defeq
	\sigma
	\left(
		\bm{\phi}_{k,j},
		\bm{F}_{k,j}
	,
		j\le i
	\right)
=
	\sigma
	\left(
		\bm{\phi}_{k,0},
		\bm{F}_{k,1}
	,\ldots
		\bm{F}_{k,i}
	\right)
$: \footnote{a.s. stands for almost surely.}
\begin{IEEEeqnarray}{rCl}
	\E 
	\left[ 
		\| \bm{F}_{k,i+1} \|^2
		|
		\mc{F}_{k,i}
	\right]
&
\le
&
	L_2 
	\left( 
		1 + \| \bm{\phi}_{k,i} \|^2
	\right)
\quad
	a.s.
\quad
	i \ge 0
,\;\;
    0 < L_2 < \infty
,\;\;
    \forall k \in \N
.
\end{IEEEeqnarray}
\end{assumption}
\begin{assumption}
\label{ass:bounded-iterates}
The iterates of \eqref{eq:joint-adaptation}--\eqref{eq:joint-combination} remain bounded:
\begin{IEEEeqnarray}{rCl}
\sup_{i}
	\left\|
		\bm{\phi}_{k,i}
	\right\|
&
<
&
	\infty
\quad
	a.s.
\quad
    \forall k \in \N
.
\end{IEEEeqnarray}
\end{assumption}

Assumption \ref{ass:connectivity-matrix} can be satisfied by letting the agents find their combination weights in a fully distributed manner (as explained in Sec. \ref{sec:multiagent-approach}).
Moreover, it can be easily relaxed such that the connectivity matrix is left stochastic; or drawn from a probability distribution, with its elements being i.i.d. random variables so that the assumption holds in expectation, i.e., $C = \E[\bm{C}]$.

Assumption \ref{ass:lipschitz-gradients} holds for very expressive function approximations, like deep neural networks with common activation functions. 

Assumption \ref{ass:stepsizes} is easily imposed in practice. The rightmost term of \eqref{eq:stepsizes} ensures that the step size for the actor goes to zero faster than the critic. This implies that the actor updates on a slower timescale than the  critic.

Assumption \ref{ass:square_integrable} is standard (and holds easily, e.g. when rewards are bounded)  and allows the noise to be averaged by the diminishing step sizes \citep[p. 12]{borkar2008stochastic}.

Assumption \ref{ass:bounded-iterates} typically requires a stability analysis \citep{Borkar99theode,borkar2008stochastic,lakshminarayanan2017stability}. One simple way to ensure this assumption would be that each agent projects its parameters to some compact set after performing the local adaptation step (e.g., by clipping the parameters).

Let us introduce the average parameter and gradient vectors:
\begin{IEEEeqnarray}{rCl}
	\overline{\bm{\phi}}_i
\defeq
	\frac{1}{N}
	\sum_{k=1}^N
		\bm{\phi}_{k,i}
,\quad
	\overline{g} (\cdot)
\defeq
	\frac{1}{N}
	\sum_{k=1}^N
		g_k(\cdot)
=
    \left(
    \begin{array}{c}
         \nabla_\xi J(\cdot) \\
         \nabla_w \widetilde{L}(\cdot)
    \end{array}
    \right)
.
\label{eq:average-parameter-gradient}
\end{IEEEeqnarray}

The following theorem shows that all agents converge to a single average parameter vector.
%
%
\begin{theorem}
Under Assumptions \ref{ass:connectivity-matrix}--\ref{ass:bounded-iterates}, each agent's parameters obtained from distributed recursion \eqref{eq:joint-adaptation}--\eqref{eq:joint-combination} converge almost surely to their average recursion:
\begin{IEEEeqnarray}{rCl}
	\lim_{i\rightarrow \infty} \bm{\phi}_{k,i}
&=&
    \lim_{i\rightarrow \infty} \overline{\bm{\phi}}_i
\quad
	a.s.
\quad
	\forall k \in \N
.
\end{IEEEeqnarray}
\label{theorem:asymptotic-agreement}
\end{theorem}
\begin{IEEEproof}
Appendix \ref{app:proofs}.2 includes the proof, which is composed of the following three steps:
First, we introduce agreement and disagreement vectors, the former being the average of all agents' parameters, and the latter being the difference between each agent's parameter and the agreement vector. 
Second, we build a recursion on the the disagreement vector and show that it converges almost surely to some point. Finally, we show that the convergent point is the null vector, implying that the agents do not deviate from their average.
\end{IEEEproof}

Our next result shows that $\overline{\bm{\phi}}_i$ converges to a stationary point of $\overline{g}$. 
To show this, we use the ordinary differential equation (ODE) method \citep{borkar2008stochastic}, which requires three additional assumptions.
Consider the following ODE:
\begin{align}
    \dot{\xi}(t) 
&=
    \nabla_\xi J(\xi(t), \pi_w(t))
\label{eq:ode-value}
,\quad
\\
    \dot{w}(t) 
&=
    \nabla_w \widetilde{L}(v_\xi(t), w(t))
.
\label{eq:ode-policy}
\end{align}
\begin{assumption}
\label{ass:martingale}
The sequence 
$
\left\{
	\bm{F}_{k,i+1}
\right\}_{i=0}^\infty
$ is a martingale difference sequence w.r.t. $\mc{F}_{k,i}$:
\begin{IEEEeqnarray}{rCl}
	\E 
	\left[ 
		\bm{F}_{k,i+1}
		|
		\mc{F}_{k,i}
	\right]
&=&
	0
\quad
	a.s.
\quad
	i \ge 0
,\;\;
    \forall k \in \N
.
\label{eq:martingale}
\end{IEEEeqnarray}
\end{assumption}
\begin{assumption}
The ODE in \eqref{eq:ode-value} has a globally asymptotically stable equilibrium $\lambda(w)$ (uniformly in $w$), where 
$
\lambda : \Re^{M_\pi} \rightarrow \Re^{M_v}
$ is a Lipschitz map.
\label{ass:ode-value}
\end{assumption}
\begin{assumption}
The ODE \eqref{eq:ode-policy} has a globally asymptotically stable attractor $\mathbb{W}^\star$.
\label{ass:ode-policy}
\end{assumption}

Although Assumptions \ref{ass:martingale}--\ref{ass:ode-value} do not hold for the deep neural network representations of the value function used for the experiments, they are convenient to simplify the analysis and rely on the two time-scale stochastic approximation analysis due to  \cite{borkar1997stochastic}.
This allows us to illustrate the behaviour of the Diff-DAC architecture with little effort. 
More general conditions can be found in, e.g., \cite{tadic2004almost,ramaswamy2017generalization, yaji2020stochastic}.

Moreover, condition \eqref{eq:martingale} in Assumption \ref{ass:martingale} would hold for a variation of Diff-A2C that used a linear critic with the so-named \textit{compatible features} condition, which is given by:
\begin{IEEEeqnarray}{rCl}
	\nabla_\xi v_{\xi_{k,i}}(s)
=
	\int_\Ac
		\nabla_w \log \pi_{w_{k,i}}(s,a)
		\ da
,
\end{IEEEeqnarray}
and ensures unbiased policy gradients even with a parametric value function approximation
\citep{Sutton1999policygradient,KondaActorCritic2003,tomczak2019compatible}.
Assumption \ref{ass:ode-value} would also hold for multiple linear critics \citep{Sutton2009,BhatnagarNaturalActorCritic2009,Scherrer2010}.

Assumption \ref{ass:ode-policy} is more general than Assumption \ref{ass:ode-value}; it holds for parametric policies for which $\nabla_w \widetilde{L}$ has multiple stationary points.

We are ready to state the main result of this section.
%
%
\begin{theorem}
Under Assumptions \ref{ass:lipschitz-gradients}--\ref{ass:ode-policy},
the average parameter converges to a stationary point of the average gradient:
\begin{IEEEeqnarray*}{rCl}
    \lim_{i\rightarrow\infty} \overline{\bm{\phi}}_i
&\in&
    \mathbb{W}^\star
\quad a.s.
\quad \text{such that}
\quad
    \overline{g}\left(\phi^\star\right) 
= 
    0
,\;
\forall \phi^\star \in \mathbb{W}^\star.
\end{IEEEeqnarray*}
\label{theorem:convergence-to-stationary}
\end{theorem}
%
%
\begin{IEEEproof}
See Appendix \ref{app:proofs}.3.
\end{IEEEproof}
%

%
\section{Numerical Experiments}
%

In this section, we illustrate the benefits of the diffusion mechanism when applied to actor-critic algorithms. First, we illustrate its stabilising effect on the sampling distribution.
Second, we show that Diff-SiAC is able to outperform Dist-MTPS.
Third, we show a regularisation effect that can lead to higher performance and better generalisation properties than the centralised architecture.
Finally, we explore the influence of the network topology and noisy links on the performance of the algorithm.

%
\subsection{Stability}
\label{ssec:stability}
%

We evaluate the stabilising properties of the Diff-DAC architecture by comparing Centralised SiAC vs Diff-SiAC on multitask variants of two classic control problems: Inverted pendulum, and Cart-pole swing-up.

The inverted pendulum consists of a rigid pole and an actuated joint with limited torque. The goal is, starting from a random position, to take the pendulum to the upright position and balance.
Cart-pole swing-up is an extension of cart-pole balance where the pole starts from the bottom and the goal is to swing the pole to the upright position and balance. 
The MRL problem consists of $25$ tasks where we vary the pole mass and length for the inverted pendulum; and the pole mass, pole  length and cart mass for Cart-pole swing-up (see task details in Appendix \ref{app:numerical-experiments}.1).

Our goal with these experiments is not to compare with state of the art centralised algorithms---which use several advancements to stabilise or improve performance---but to evaluate whether diffusion is an effective mechanism to stabilise learning.
Thus, we compare Centralised SiAC vs. Diff-SiAC, which are two versions of the same algorithm, described in Sec. \ref{ssec:vac}, with the same advantage estimates and the same neural network architectures
(agent details in Appendix \ref{app:numerical-experiments}.1).

For Diff-SiAC, the network consists of $N=25$ agents (each agent is allocated one task), 
in a random strongly connected graph with average degree $|\overline{\N}| \defeq \sum_{k=1}^N |\N_k| \approx 4.2$.
The connectivity matrix $C$ was obtained with the Hastings-rule \cite[p.492]{sayed2014adaptation}, 
so that \eqref{eq:non-negative-coefficients}--\eqref{eq:aperiodic} hold.
The random network topology aims to reflect the sparse connectivity that appears naturally when agents and data are geographically distributed---it is not related to any form of task similarity.

\begin{figure}
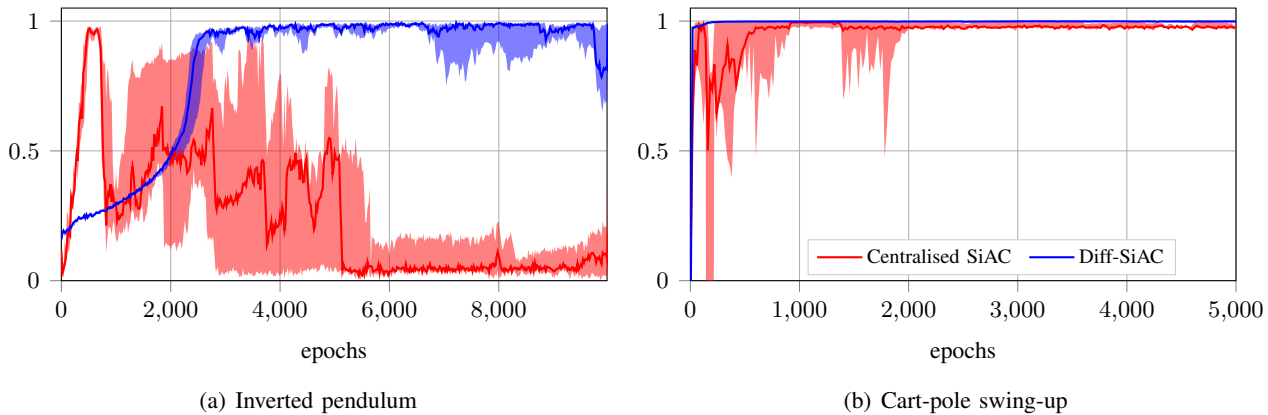

\subfigure[Inverted pendulum]{\centering \input{pendulum_multitask_25_4_connectivity.tikz}}
\subfigure[Cart-pole swing-up]{\centering \input{cartpoleswingup_multitask_25_4_connectivity.tikz}}
\caption{Stability. Learning curves from training with 25 randomly sampled tasks of Inverted pendulum (left) and Cart-pole swing-up (right). Plots show median and first and third quartiles of the return averaged over all tasks, and 6 seeds. 
Centralised SiAC (red) is unstable, while Diff-SiAC (blue) learns robustly.}
\label{fig:inverted_pendulum-swing_up}
\end{figure}

The results are reported as the return for all tasks, which is averaged over $10$ test episodes per epoch and scaled by the maximum value achieved by any algorithm.
Each epoch consists of $5$ episodes per agent in Diff-SiAC, 
and $5N$ episodes in total for Centralised SiAC, 
so that both algorithms simulate the same number of episodes.
Each experiment was repeated with $6$ seeds.
Figure \ref{fig:inverted_pendulum-swing_up} shows the median and first and third quartiles of the distribution of the return.

We observe that Centralised SiAC starts learning faster than Diff-SiAC in multitask inverted pendulum (left), and achieves near maximum performance for both problems. 
Faster learning under the centralised architecture is expected since it can compute the gradients with data from all tasks at every iteration, while Diff-SiAC has to wait until the parameters are diffused across the network. 
However, Centralised SiAC is unstable in both problems, which is also expected since the samples used to estimate the gradients are highly correlated and Centralised SiAC has no decorrelation mechanism.

Diff-SiAC converges more slowly but is much more stable in both problems.
Recall that Centralised SiAC and Diff-SiAC have the same network architecture, sample from multiple environments in parallel and are trained with the same amount of data in total.
We therefore believe that this enhanced stability is due to a bias removal effect on the sampling distribution introduced by the diffusion mechanism, similar to the distributed exploration capabilities exhibited when applying diffusion to off-policy evaluation \citep{valcarcel2013distributed}.

We performed similar experiments for the single task problem---where Centralised SiAC learns from multiple copies of the same task, and all Diff-SiAC agents train on copies of the same task---and observed the same stability improvement (see results in Appendix \ref{app:singletask}).

The instability of Centralised SiAC might be alleviated in multiple ways, like adding a \textit{replay buffer} \citep{mnih2013playing,Lillicrap2015Continuous}. 
Our goal with these experiments is precisely to show that diffusion can be used as an alternative stabilisation technique.

%
\subsection{Comparison with Dist-MTPS}
\label{ssec:comparison-dist-mtls}
%

We compare Diff-SiAC with Dist-MTPS on an MRL extension of the standard Cart-pole balance problem, which is the only environment considered in this paper that is controllable with linear policies learnt from raw data.
The agents have to learn $N=25$ tasks, characterised by different pole masses and lengths 
(see Appendix \ref{app:numerical-experiments}.2 for details).

In particular, we compare against two variants of Dist-MTPS, which vary by their use of differing policy search methods: REINFORCE \citep{williams1992simple} and PoWER \citep{kober2009policy}.

\begin{figure}
\begin{minipage}[t]{0.5\textwidth}
	\vspace{0pt}
    \centering \input{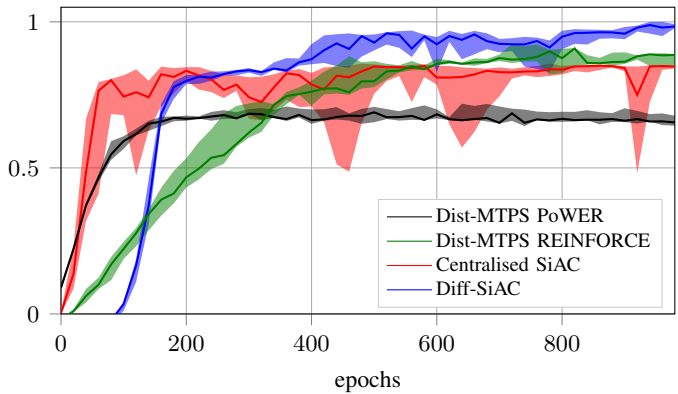}
\end{minipage}\hfill
\begin{minipage}[t]{0.43\textwidth}
	\vspace{0pt}\raggedright
	\caption{Multitask Cart-pole balance with continuous action space.
	Centralised SiAC learns faster than the distributed approaches and achieves similar performance to Dist-MTPS REINFORCE, but Diff-SiAC achieves the best asymptotic performance and with lower variance.
    Each epoch consists of $5$ episodes per agent in Diff-SiAC, and $5N$ episodes for Dist-MTPS and Centralised SiAC, so that all algorithms run the same number of episodes.
    Results are averaged over tasks and $5$ seeds.}
\label{fig:mrl-cart-pole-balance}
\end{minipage}
\end{figure}

Figure \ref{fig:mrl-cart-pole-balance} shows that Diff-SiAC learns faster and reaches higher return than Dist-MTPS REINFORCE.
Dist-MTPS PoWER converges faster than Diff-SiAC, however the return of the latter is significantly higher.
This is remarkable since Dist-MTPS learns a different policy for each task, while Diff-DAC agents learn a single policy for all tasks.

We also compare with Centralised SiAC in this example. 
Consistent with the stability experiments, Figure \ref{fig:mrl-cart-pole-balance} shows that Diff-SiAC learns more slowly but with lower variance than Centralised SiAC. 
In addition, we observe a new effect: Diff-SiAC clearly achieves higher performance than the centralised architecture.
This is a very interesting feature of the diffusion mechanism that has been already reported for non-convex optimisation problems \citep[Ch. 4]{valcarcel2017phdthesis}.
We believe this is due to a new form of regularisation, which we explore further in the next subsection.

%
\subsection{Regularisation}
\label{ssec:regularisation}
%

We extend our analysis to multitask variants of two more challenging and higher dimensional environments: Acrobot and MuJoCo Hopper.

In Acrobot the agent must learn a policy to swing a two-link robot from its initial position, approximately hanging down, to a goal position of balancing straight upwards. 
The randomisation of the environment dynamics generates tasks by sampling the length, mass and inertia of the robot uniformly from a distribution centred on the values of the standard Acrobot environment. This formulation was originally introduced by \cite{PackerGao:1810.12282}. We design a MRL problem by sampling 25 tasks (see Appendix \ref{app:numerical-experiments}.3 for details).

The objective in MuJoCo Hopper is to learn a policy to keep the agent from falling over and to enable the agent to move as quickly as possible in any direction.
We implement two MRL problems, consisting of 25 tasks each.  
In one set of experiments, we change the mass of the agent for each task. 
The second set of experiments introduces wind. We change the direction and intensity of the wind for each task (see Appendix \ref{app:numerical-experiments}.3 for details).

The randomisation of the environment parameters---pole mass, pole length and inertia for Acrobot and agent mass and wind speed and direction for Hopper---is calibrated to generate an MRL problem more challenging than the original single-task environments. 

For these MRL problems, we compare three agent implementations: 
\textit{i)} Diff-A2C on a network of $N=25$ agents, each one with a single process sampling from one of the tasks;
\textit{ii)} Centralised A2C where one agent has $25$ processes, one per task, so it simultaneously learns from experience collected on all tasks;
and \textit{iii)} Specialised A2C agents which train on each of the tasks individually but each one with the equivalent resources to the Centralised A2C agent (i.e., each of the 25 specialised agents is assigned $25$ copies of one of the sampled environments).

The comparison of Diff-A2C and Centralised A2C enables us to compare two learning architectures using approximately the same amount of computation. Centralised A2C centralises learning while maintaining decentralised sampling from environments run in parallel. Diff-A2C decentralises both sampling and learning. 
The specialised agents provide an upper bound to performance during training thereby offering an opportunity to evaluate the performance gap incurred by the multitask agents.

For this set of experiments, all agents are trained using the RMSProp optimiser \citep{tieleman2012lecture}. This provides an opportunity to show that the Diff-DAC architecture is optimiser agnostic in practice--although it is derived and analysed assuming vanilla stochastic gradient descent. Indeed, we attain similar results when running the experiments presented here with the Adam optimiser. Full details of agent hyperparameters are provided in Appendix \ref{app:numerical-experiments}.

The ring topology is used for Diff-A2C in this section. This topology can be challenging due to the sparsity of the connectivity, which implies a slower diffusion rate (it takes up to 12 iterations to diffuse each agent's parameters to every other agent) and less robust communication (there are only two possible paths between every pair of agents).

\begin{figure}
\begin{minipage}[t]{0.5\textwidth}
	\vspace{0pt}
	{\centering \input{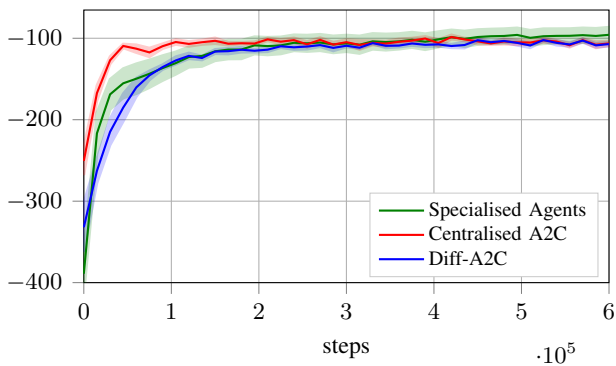}}
\end{minipage}
\begin{minipage}[t]{0.43\textwidth}
    \vspace{0pt}\raggedright
	\caption{Regularisation Acrobot. Learning curves from training with 25 randomly sampled tasks of Random Extreme Acrobot. For each experiment, we average performance over all 25 tasks. The plot shows the average $\pm$ 1 standard error over 8 seeds.
	Centralised A2C learns faster. Diff-A2C achieves similar performance to Centralised A2C. Specialised agents achieve the highest performance, but with small gap.}
\label{fig:acrobot-learning}
\end{minipage}
\end{figure}

\begin{figure}
    \subfigure[Random Mass MuJoCo Hopper] {\centering\input{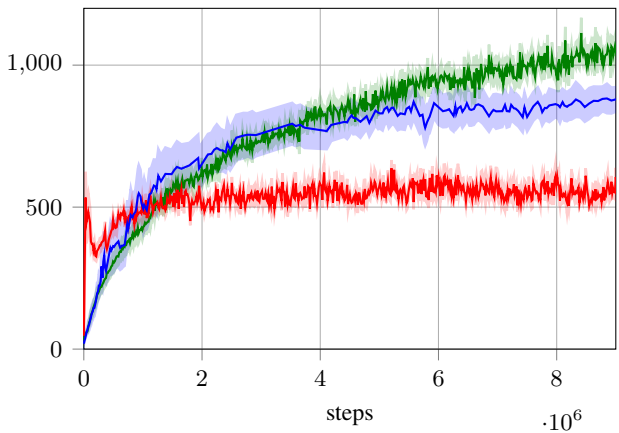}}
    \subfigure[Random Wind MuJoCo Hopper] {\centering\input{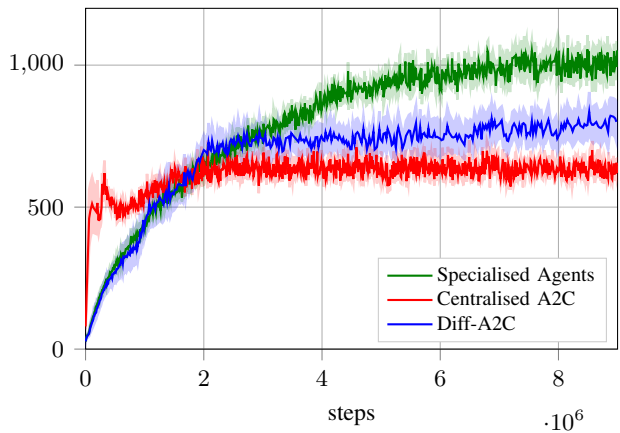}}
    \caption{Regularisation MuJoCo Hopper. Learning curves from training with 25 randomly sampled tasks of Random Mass (left) and Random Wind (right) MuJoCo Hopper. For each experiment, we average performance over all 25 tasks. The plots show the average $\pm$ 1 standard error (of averages) over 8 seeds.
    Specialised agents achieve the highest performance, with a significant gap w.r.t. to the other agents. Diff-A2C agents clearly outperform Centralised A2C.}
\label{fig:random-hopper-learning}
\end{figure}

Figure \ref{fig:acrobot-learning} displays the results for experiments on the Acrobot environments. Diff-A2C and Centralised A2C achieve the same final performance in training while the specialised agents achieve slightly greater ultimate performance. The optimality gap is  small, meaning that both Diff-A2C and Centralised A2C learn a single policy that is expressive enough to control the 25 tasks near optimally.
Consistent with previous experiments, we see that Centralised A2C learns faster initially. 
In this case however, Centralised A2C does not exhibit stability issues, as A2C's decentralised sampling stabilises the sampling distribution, similar to the stabilising effect of the Diff-DAC architecture.

Figure \ref{fig:random-hopper-learning} shows results for the two MRL problems with MuJoCo Hopper: random mass and random wind.
These are significantly more challenging tasks than Acrobot.
We observe the pattern of initially fast learning for the centralised architecture.
But now we can clearly see that the optimality gap is smaller for Diff-A2C in both set of experiments.
This supports the hypothesis that diffusion induces some form of regularisation effect that allows the agents to find a better policy that can control the set of tasks more effectively.

However, since each Diff-A2C agent maintains its own policy, it is not clear whether the performance improvement over the centralised architecture is due to finding a better policy or to being able to exploit policies specialised to each task.
In order to investigate, we look at how similar the policies learnt by the Diff-A2C agents are in terms of parameters and behaviour on the tasks used in training.
If the policies of all Diff-A2C agents are equal, then we can be sure that the policy found via Diff-A2C is better than that found by Centralised A2C.

\begin{table}
\centering
\begin{tabular}{lcc}
\toprule
                  & Diff-A2C & Specialised Agents \\
\midrule
Randomised Acrobot & 3.56\%      & 59.27\%           \\
Random Mass Hopper & 3.10\%      & 114.08\%           \\
Random Wind Hopper & 3.14\%      & 116.23\%           \\
\bottomrule
\end{tabular}
\vspace*{0.2cm}
\caption{Parameter similarity. Relative deviation of final parameters of trained agents. This is determined by calculating the mean and the standard deviation of each parameter of the trained agent over 25 agents (one per task). The standard deviations are then divided by the mean. The reported values are the average over the parameters and over 8 seeds.}
\label{tab:param-variance}
\end{table}
\begin{table}
\centering
\begin{tabular}{lccc}
\toprule
                   & \wrap{Training\\Environment}     & \wrap{Out-of-Sample\\Easy Environment} & \wrap{Out-of-Sample\\Hard Environment} \\
\midrule
Specialised A2C    & \textbf{-82.1}  (-98.3,  -66.0)    & -80.0 (-79.1, -62.8)   & -163.8 (-182.3, -145.3)  \\
Diff-A2C           & -109.4 (-128.0, -90.9)   & \textbf{-48.8} (-57.4, -40.2)   & \textbf{-122.6} (-142.9, -102.3)  \\
Centralised A2C    & -115.6 (-138.6, -92.6)   & -53.4 (-63.6, -43.3)   & -141.8 (-166.2, -117.3)  \\
\bottomrule
\end{tabular}
\vspace*{0.2cm}
\caption{Zero-shot learning. Evaluation performance for trained agents on unseen Acrobot tasks. The values are averages over 8 seeds and 10 Episodes each. For the Specialised and Diff-A2C agents, a single agent from each experiment was chosen at random and evaluated on the environment they were trained on as well as the same previously unseen easy and hard environments. Values in brackets denote a 95\% confidence interval.}
\label{tab:acrobot-generalisation}
\end{table}

Table \ref{tab:param-variance}, shows the relative deviation of the parameters as a measure of the similarity of the agents at the end of training. 
For Diff-A2C and specialised agents, the value is calculated by dividing the standard deviation by the mean of each model parameter across all agents, then averaging the results over all parameters, and finally averaging over 8 runs with differing random seeds.
The results show that all Diff-A2C agents converge to very similar policy networks. 
On the other hand, the specialised agents converge to policy networks that are significantly different from each other.
This is indeed consistent with Theorem \ref{theorem:asymptotic-agreement}, which shows almost certain asymptotic agreement among the agents,
and supports the hypothesis that diffusion allows us to find a single policy that performs better than the one found by the centralised architecture.

In order to further validate the convergence of Diff-A2C agents to a common solution, we ran each of the 25 trained agents on the task they were trained on as well as the tasks of all other agents in the network. On average, agents achieved 0.07\% ($\pm$ 3.4) higher return, over 10 episodes, in the task they were trained on compared to the tasks of their peers (result averaged over all agents and 8 random seeds). Since the performance difference is not statistically significant, we conclude the small deviations in the policy parameters are not significant.

The above analysis suggests that the policy learnt by diffusion should be better able to generalise across tasks than the one found by the centralised architecture. 
To consider the question of generalisation, we performed a zero-shot learning experiment in the Acrobot environment, evaluating our trained agents on newly sampled environments (details of these environments are provided in Appendix \ref{app:numerical-experiments}.4). 
Results are presented in Table \ref{tab:acrobot-generalisation}, and show that Diff-A2C yields agents better able to generalise to new environments. The Centralised A2C agents show some ability to generalise to new environments but achieve lower returns in the evaluation when compared to Diff-A2C despite attaining near identical performance at the end of training. Unsurprisingly, the specialised agents do not generalise well. Specialised agents attain returns approximately 33\% higher than Diff-A2C agents on the environment in which they specialise but are outperformed by 31\% and 25\% when transferred to held out easy and hard environments respectively.\footnote{Table \ref{tab:acrobot-generalisation-relative} in Appendix \ref{app:numerical-experiments}.4 presents the results in terms of performance relative to specialised agents in full. Relative performance values are calculated from Table \ref{tab:acrobot-generalisation}.}

%
\subsection{Network dependence}
\label{ssec:network-dependence}
%

In this section, we perform experiments to study the influence of the communications network on the performance of the proposed Diff-DAC architecture. 
First, we evaluate the impact of the network topology by varying the size of the graph and its neighbourhoods. 
We then explore the impact of noisy or intermittent links on agent performance.

We evaluate the impact of the network topology with Diff-SiAC agents learning the \textit{single-task} cart-pole balance problem, where all agents operate in different copies of exactly the same environment (see details in Appendix \ref{app:numerical-experiments}.4). 
Figure \ref{fig:connectivity} shows that for the same network size, $N=25$, 
a relatively sparse network, $|\overline{\N}| \approx N/6$, learns slightly more slowly but achieves performance similar to a more dense network, $|\overline{\N}| \approx N/3$.
This is consistent with the theoretical analysis that shows that as long as the connectivity matrix, $C$, satisfies conditions \eqref{eq:non-negative-coefficients}--\eqref{eq:aperiodic}, 
asymptotic performance is guaranteed independent on the network sparsity.
From a practical point of view, this is a promising result that suggests that communication and computational costs per agent can be reduced without penalising performance, since they scale with the number of neighbours.
In addition, we see that larger number of agents $N=100$ improves the final performance, suggesting that the regularisation effect discussed in Section \ref{ssec:regularisation} is stronger with a larger number of agents.

\begin{figure}
\begin{minipage}[t]{0.5\textwidth}
	\vspace{0pt}
    \centering \input{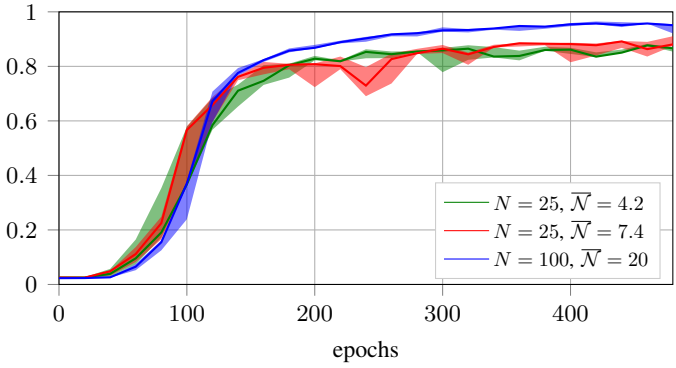}
\end{minipage}\hfill
\begin{minipage}[t]{0.46\textwidth}
	\vspace{0pt}\raggedright
    \caption{Network topology. Influence of network topology in single-task cart-pole balance with continuous action space. Diffusion is able to combine the experience of all agents in a way that is relatively insensitive to network sparsity. Moreover, by increasing the number of agents, we obtain higher performance, suggesting Diff-SiAC is able to find a better local optimum. Results are averaged over all agents and 5 seeds.}
\label{fig:connectivity}
\end{minipage}
\end{figure}

\begin{figure}
    \centering
    \subfigure[Acrobot]{\centering \input{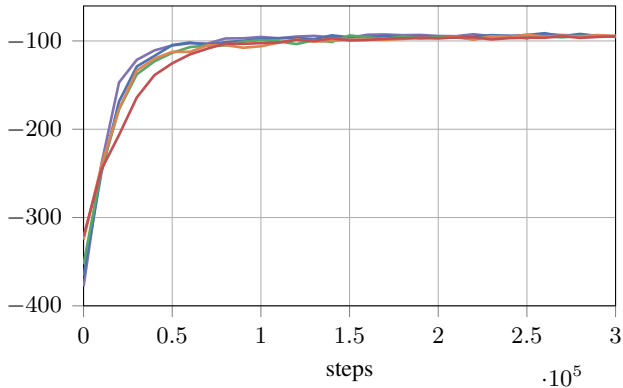}}
    \subfigure[MuJoCo Hopper] {\centering\input{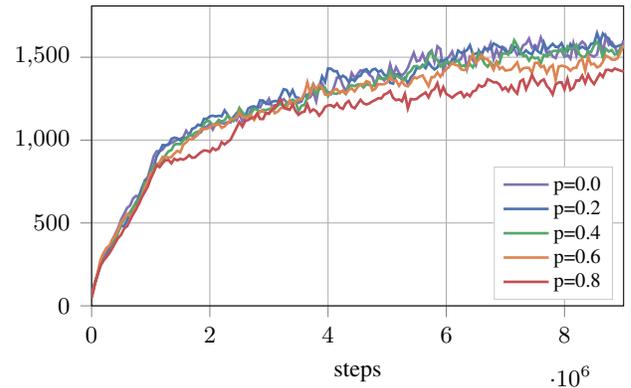}}
    \caption{\textbf{(Noisy links)} Learning trajectories of Diff-A2C agents with differing probabilities of links being dropped at an update step. The probability, denoted by $p$, parameterises independent Bernoulli distributions, one per connection. Experiments conducted with 14 agents connected in a ring topology. The plots show performance averaged over all agents in the graph and averaged over 8 seeds.}
\label{fig:dropped-links}
\end{figure}

Finally, we consider the impact of failing links on training in the original (single-task) Acrobot and MuJoCo Hopper environments.
These experiments emulate issues of noisy or interrupted message passing, typically seen in real deployments. 
As shown in Figure \ref{fig:dropped-links}, as the probability of any link in the network failing at a given update step increases, the Diff-A2C agents' learning is slowed. However, the agents converge towards similar final performance. This shows that diffusion is robust to failed links even when the failure rate is as high as 80\%.
This was predicted in Sec. \ref{sec:convergence-analysis}, where we discuss that conditions on the connectivity matrix need only hold in expectation. 
The Diff-A2C experiments were conducted with a ring topology meaning that dropped links easily cause the graph to become disconnected. We would therefore expect this topology to be particularly vulnerable to dropped links. In the case of a centralised agent, however, should a single environment instance fail, the agent's learning is liable to fail completely unless explicit countermeasures are introduced. 

%
\section{Conclusions}
\label{sec:conclusions}
%

We considered the MRL problem where tasks are MDPs with parametrised dynamics where parameters are drawn from some distribution.

We used standard optimal control and convex optimisation theory to derive a tabular actor-critic algorithm as an instance of dual ascent for finding the saddle point of a   Lagrangian. 
We also re-derived the policy gradient theorem from this same formulation.
These derivations are interesting in themselves and provide novel insights in the actor-critic framework.
By approximating the exact method with parametric approximations, we obtained the Diff-DAC architecture, which can be applied to transform many actor-critic algorithms into fully-distributed implementations that scale to a large number of tasks. 
We also proved convergence of Diff-DAC to a common policy under mild assumptions, and to stationary point of the joint value and policy gradients under more restrictive assumptions.

Results from our experiments showed that the Diff-DAC architecture can achieve higher performance than Dist-MTPS.
This is a remarkable result since the Diff-DAC agents converge to a single policy that behaves better than the task-dependent policies obtained by Dist-MTPS.
Moreover, Diff-DAC can solve complex problems that are uncontrollable from raw data by linear policies, while Dist-MTPS requires feature engineering.

We also demonstrate that Diff-DAC is very stable and achieves similar or higher performance and shows better generalisation than a comparable centralised approach.
This suggests that the sparse connectivity among agents induces a regularisation effect that helps them to achieve better local optima.

Finally, the experiments with intermittent connections between agents provide empirical evidence of the robustness to communication disruption Diff-DAC achieves in distributed learning systems.

\section*{Acknowledgements}
%
We thank David Baldazo, Daniel Garc\'ia-Oca{\~n}a Hern\'andez, and Santiago Zazo for insightful preliminary discussions; Felix Leibfried for his support with the experiments; Peter Vrancx, Haitham Bou-Ammar and Rasul Tutunov for helpful comments; and the anonymous reviewers for their comments and suggestions that have helped to improve the presentation of the paper.


\bibliographystyle{agsm}
\bibliography{myreferences,myarticles}

@inproceedings{tomczak2019compatible,
  title={Compatible features for Monotonic Policy Improvement},
  author={Tomczak, Marcin B and Valcarcel Macua, Sergio and de Cote, Enrique Munoz and and Vrancx, Peter},
  journal={{NeurIPS} Optimization Foundations for Reinforcement Learning Workshop},
  year={2019}
}

@ARTICLE{valcarcel2013distributed, 
author={Valcarcel Macua, Sergio and J. Chen and S. Zazo and A. H. Sayed}, 
journal={{IEEE} Transactions on Automatic Control}, 
title={Distributed Policy Evaluation Under Multiple Behavior Strategies}, 
year={2015}, 
volume={60}, 
number={5}, 
pages={1260-1274}, 
month={May}
}

@phdthesis{valcarcel2017phdthesis,
    title    = {Distributed optimization, control and learning in multiagent networks},
    school   = {Universidad Polit\'ecnica de Madrid},
    author   = {Valcarcel Macua, Sergio},
    year     = {2017}, 
}

@article{valcarcel2017diffdac,
  title={Diff-{DAC}: Distributed actor-critic for average multitask deep reinforcement learning},
  author={Valcarcel Macua, Sergio and Tukiainen, Aleksi and Hern{\'a}ndez, Daniel Garc{\'\i}a-Oca{\~n}a and Baldazo, David and de Cote, Enrique Munoz and Zazo, Santiago},
  journal={arXiv preprint arXiv:1710.10363},
  year={2017}
}

@InProceedings{andreas2017Modular,
  title = 	 {Modular Multitask Reinforcement Learning with Policy Sketches},
  author = 	 {Jacob Andreas and Dan Klein and Sergey Levine},
  booktitle = {Proc. Int. Conf. on Machine Learning (ICML)},
  pages = 	 {166--175},
  year = 	 {2017},
  month = 	 {Aug},
}

@Book{arrow1958studies,
  Title                    = {Studies in Linear and Non-linear Programming},
  Author                   = {Arrow, K. J. and Hurwicz, L. and Uzawa, H.},
  Publisher                = {Stanford University Press},
  Year                     = {1958},

  Lccn                     = {58013532}
}

@inproceedings{assran2019gossip,
  title={Gossip-based Actor-Learner Architectures for Deep Reinforcement Learning},
  author={Assran, Mahmoud and Romoff, Joshua and Ballas, Nicolas and Pineau, Joelle and Rabbat, Michael},
  booktitle={Advances in Neural Information Processing Systems (NIPS)},
  pages={13320--13330},
  year={2019}
}

@techreport{baird1993advantage,
  title={Advantage updating},
  author={Baird III, Leemon C},
  year={1993},
  institution={Wright Lab Wright-Patterson AFB OH}
}

@Book{bertsekas2009convex,
  Title                    = {Convex Optimization Theory},
  Author                   = {Bertsekas, D. P.},
  Publisher                = {Athena Scientific},
  Year                     = {2009}
}

@Book{Bertsekas2012DPvol2,
  Title                    = {Dynamic Programming and Optimal Control},
  Author                   = {Bertsekas, D. P.},
  Publisher                = {Athena Scientific},
  Year                     = {2012},
  Edition                  = {4th},
  Volume                   = {2}
}

@Article{BhatnagarNaturalActorCritic2009,
  Title                    = {Natural Actor-critic Algorithms},
  Author                   = {Bhatnagar, S. and Sutton, R. S. and Ghavamzadeh, M. and Lee, M.},
  Journal                  = {Automatica},
  Year                     = {2009},

  Month                    = nov,
  Number                   = {11},
  Pages                    = {2471--2482},
  Volume                   = {45},

  ISSN                     = {0005-1098},
  Issue_date               = {November, 2009},
  Numpages                 = {12},
  Publisher                = {Elsevier}
}

@Article{bianchi_convergence_2013,
  Title                    = {Convergence of a Multi-Agent Projected Stochastic Gradient Algorithm for Non-Convex Optimization},
  Author                   = {Bianchi, P. and Jakubowicz, J.},
  Journal                  = {IEEE Transactions on Automatic Control},
  Year                     = {2013},

  Month                    = {Feb},
  Number                   = {2},
  Pages                    = {391-405},
  Volume                   = {58}
}

@article{borkar1997stochastic,
title = "Stochastic approximation with two time scales",
journal = "Systems and Control Letters",
volume = "29",
number = "5",
pages = "291 - 294",
year = "1997",
author = "Vivek S. Borkar"
}

@Article{Borkar99theode,
  Title                    = {{The O.D.E. method for convergence of stochastic approximation and reinforcement learning}},
  Author                   = {V. S. Borkar and S.P. Meyn},
  Journal                  = {SIAM Journal on Control and Optimization},
  Year                     = {1999},
  Pages                    = {447-469},
  Volume                   = {38}
}

@Book{borkar2008stochastic,
  Title                    = {Stochastic Approximation: A Dynamical Systems Viewpoint},
  Author                   = {Borkar, V.S.},
  Publisher                = {Cambridge University Press},
  Year                     = {2008}
}

@inproceedings{bou-ammar2014online,
  title={Online multi-task learning for policy gradient methods},
  author={Bou-Ammar, H. and Eaton, E. and Ruvolo, P. and Taylor, M.},
  booktitle={Proc. Int. Conf. on Machine Learning (ICML)},
  pages={1206--1214},
  year={2014}
}

@Book{boyd2004convex,
  Title                    = {Convex Optimization},
  Author                   = {Boyd, S.P. and Vandenberghe, L.},
  Publisher                = {Cambridge University Press},
  Year                     = {2004}
}

@Article{Chen2013a_distributed,
  Title                    = {Distributed {Pareto} Optimization via Diffusion Strategies},
  Author                   = {Chen, J. and Sayed, A. H.},
  Journal                  = {{IEEE} Journal of Selected Topics in Signal Processing},
  Year                     = {2013},

  Month                    = apr,
  Number                   = {2},
  Pages                    = {205--220},
  Volume                   = {7},

  Doi                      = {10.1109/JSTSP.2013.2246763},
  ISSN                     = {1932-4553, 1941-0484},
  Urldate                  = {2013-09-11}
}

@inproceedings{el2017scalable,
  title={Scalable Multitask Policy Gradient Reinforcement Learning.},
  author={El Bsat, S. and Bou-Ammar, H. and Taylor, M. E.},
  booktitle={AAAI Conf. on Artificial Intelligence (AAAI)},
  pages={1847--1853},
  year={2017}
}

@inproceedings{espeholt2018impala,
  title={IMPALA: Scalable Distributed Deep-RL with Importance Weighted Actor-Learner Architectures},
  author={Espeholt, Lasse and Soyer, Hubert and Munos, R{\'e}mi and Simonyan, Karen and Mnih, Volodymyr and Ward, Tom and Doron, Yotam and Firoiu, Vlad and Harley, Tim and Dunning, Iain and others},
  booktitle={ICML},
  year={2018}
}

@inproceedings{fu2016one,
  title={One-shot learning of manipulation skills with online dynamics adaptation and neural network priors. In 2016 IEEE},
  author={Fu, Justin and Levine, Sergey and Abbeel, Pieter},
  booktitle={{IEEE} RSJ Int. Conf. on Intelligent Robots and Systems ({IROS})},
  year=2016,
  pages={4019--4026}
}

@Misc{gartner2015,
  Title                    = {Gartner Says 6.4 Billion Connected "Things" Will Be in Use in 2016, Up 30 Percent From 2015},
  Author                   = {R. van der Meulen},
  HowPublished             = {\url{http://www.gartner.com/newsroom/id/3165317}},
  Month                    = {Nov},
  Year                     = {2015}
}

@book{golub1996matrix,
  title={Matrix Computations},
  author={Golub, G.H. and Van Loan, C.F.},
  year={1996},
  publisher={Johns Hopkins University Press}
}

@Article{GrondmanActorCritic2012,
  Title                    = {A Survey of Actor-Critic Reinforcement Learning: Standard and Natural Policy Gradients},
  Author                   = {I. Grondman and L. Busoniu and G. A. D. Lopes and R. Babuska},
  Journal                  = {IEEE Transactions on Systems, Man, and Cybernetics, Part C (Applications and Reviews)},
  Year                     = {2012},

  Month                    = {Nov},
  Number                   = {6},
  Pages                    = {1291-1307},
  Volume                   = {42}
}

@InCollection{heess2015learning,
  Title                    = {Learning Continuous Control Policies by Stochastic Value Gradients},
  Author                   = {Heess, Nicolas and Wayne, Gregory and Silver, David and Lillicrap, Tim and Erez, Tom and Tassa, Yuval},
  Booktitle                = {Advances in Neural Information Processing Systems (NIPS)},
  Year                     = {2015},
  Pages                    = {2926--2934}
}

@Book{horn1990matrix,
  Title                    = {Matrix Analysis},
  Author                   = {Horn, R.A. and Johnson, C.R.},
  Publisher                = {Cambridge University Press},
  Year                     = {1990},

  ISBN                     = {9780521386326},
  Lccn                     = {lc85007736}
}

@Article{Kar2012,
  Title                    = {{QD}-Learning: A Collaborative Distributed Strategy for Multi-Agent Reinforcement Learning Through Consensus + Innovations},
  Author                   = {S. Kar and J. M. F. Moura and H. V. Poor},
  Journal                  = {IEEE Transactions on Signal Processing},
  Year                     = {2013},
  Number                   = {7},
  Pages                    = {1848-1862},
  Volume                   = {61},

  ISSN                     = {1053-587X}
}

@incollection{karp1972reducibility,
  title={Reducibility among combinatorial problems},
  author={Karp, R. M.},
  booktitle={Complexity of computer computations},
  pages={85--103},
  year={1972},
  publisher={Springer}
}

@inproceedings{kingma2015adam,
  title={Adam: A method for stochastic optimization},
  author={Kingma, DP and Ba, J. L.},
  booktitle={Proc. Int. Conf. on Learning Representations (ICLR)},
  year={2015}
}

@inproceedings{kober2009policy,
  title={Policy search for motor primitives in robotics},
  author={Kober, J. and Peters, J. R.},
  booktitle={Advances in Neural Information Processing Systems (NIPS)},
  pages={849--856},
  year={2009}
}

@Article{KondaActorCritic2003,
  Title                    = {On Actor-Critic Algorithms},
  Author                   = {Konda, Vijay R. and Tsitsiklis, John N.},
  Journal                  = {SIAM Journal on Control and Optimization},
  Year                     = {2003},
  Month                    = apr,
  Number                   = {4},
  Pages                    = {1143--1166},
  Volume                   = {42},
  Publisher                = {Society for Industrial and Applied Mathematics}
}

@article{lakshminarayanan2017stability,
  title={A stability criterion for two timescale stochastic approximation schemes},
  author={Lakshminarayanan, Chandrashekar and Bhatnagar, Shalabh},
  journal={Automatica},
  volume={79},
  pages={108--114},
  year={2017},
  publisher={Elsevier}
}

@Unpublished{Lillicrap2015Continuous,
  Title                    = {Continuous control with deep reinforcement learning},
  Author                   = {Lillicrap, T. P. and Hunt, J. J. and Pritzel, A. and Heess, N. and Erez, T. and Tassa, Y. and Silver, D. and Wierstra, D.},
  journal={arXiv preprint arXiv:1509.02971v1},
  Year                     = {2015}
}

@incollection{MeloFitted2008,
	title={Fitted Natural Actor-Critic: A New Algorithm for Continuous State-Action {MDP}s},
	author={Melo, F. S. and Lopes, M.},
	year={2008},
	booktitle={Machine Learning and Knowledge Discovery in Databases},
	volume={5212},
	publisher={Springer},
	pages={66-81}
}

@article{mnih2013playing,
  title={Playing atari with deep reinforcement learning},
  author={Mnih, V. and Kavukcuoglu, K. and Silver, D. and Graves, A. and Antonoglou, I. and Wierstra, D. and Riedmiller, M.},
  journal={arXiv preprint arXiv:1312.5602},
  year={2013}
}

@inproceedings{mnih2016asynchronous,
  title={Asynchronous methods for deep reinforcement learning},
  author={Mnih, V. and Badia, A. Puigdomenech and Mirza, M. and Graves, A. and Lillicrap, T. and Harley, T. and Silver, D. and Kavukcuoglu, K.},
  booktitle={International Conference on Machine Learning},
  pages={1928--1937},
  year={2016}
}

@InProceedings{ng1999policy,
  Title                    = {Policy Search via Density Estimation.},
  Author                   = {Ng, Andrew Y and Parr, Ronald and Koller, Daphne},
  Booktitle                = {Advances in Neural Information Processing Systems (NIPS)},
  Year                     = {1999},
  Pages                    = {1022--1028},
  Note 					   = {Denver, CO, USA, Dec. 1999},
}

@misc{OpenAI2016gym,
	Author = {Brockman, G. and Cheung, V. and Pettersson, L. and Schneider, J. and Schulman, J. and Tang, J. and Zaremba, W.},
    Title = {{OpenAI Gym}},
	Year = {2016},
	Eprint = {arXiv:1606.01540},
}

@Article{Powell2011,
  Title                    = {A review of stochastic algorithms with continuous value function approximation and some new approximate policy iteration algorithms for multidimensional continuous applications},
  Author                   = {Powell, Warren B. and Ma, Jun},
  Journal                  = {Journal of Control Theory and Applications},
  Year                     = {2011},
  Number                   = {3},
  Pages                    = {336--352},
  Volume                   = {9}
}

@Book{Puterman2005,
  Title                    = {{Markov Decision Processes: Discrete Stochastic Dynamic Programming}},
  Author                   = {Puterman, M. L.},
  Edition				   = {2nd},
  Publisher                = {John Wiley \& Sons},
  Year                     = {2005},
}

@article{ramaswamy2017generalization,
  title={A generalization of the {B}orkar-{M}eyn theorem for stochastic recursive inclusions},
  author={Ramaswamy, Arunselvan and Bhatnagar, Shalabh},
  journal={Mathematics of Operations Research},
  volume={42},
  number={3},
  pages={648--661},
  year={2017},
  publisher={INFORMS}
}

@Article{sayed2014adaptation,
  Title                    = {Adaptation, learning, and optimization over networks},
  Author                   = {Sayed, A. H.},
  Journal                  = {Foundations and Trends in Machine Learning},
  Year                     = {2014},
  Number                   = {4-5},
  Pages                    = {311--801},
  Volume                   = {7}
}

@InCollection{Scherrer2010,
  Title                    = {Should one compute the Temporal Difference fix point or minimize the {Bellman} Residual? {The} unified oblique projection view},
  Author                   = {B. Scherrer},
  Booktitle                = {Proc. Int. Conf. on Machine Learning (ICML)},
  Year                     = {2010},
  Pages                    = {959--966}
}

@article{schulman2015high,
  title={High-dimensional continuous control using generalized advantage estimation},
  author={Schulman, John and Moritz, Philipp and Levine, Sergey and Jordan, Michael and Abbeel, Pieter},
  journal={arXiv preprint arXiv:1506.02438},
  year={2015}
}

@INPROCEEDINGS{Sutton1999policygradient,
    author = {R. S. Sutton and D. Mcallester and S. Singh and Y. Mansour},
    title = {Policy gradient methods for reinforcement learning with function approximation},
    booktitle = {Advances in Neural Information Processing Systems (NIPS)},
    year = {1999},
    pages = {1057--1063},    
}

@InCollection{Sutton2009,
  Title                    = {{Fast gradient-descent methods for temporal-difference learning with linear function approximation}},
  Author                   = {Sutton, R. S. and Maei, H. R. and Precup, D. and Bhatnagar, S. and Silver, D. and Szepesvari, C. and Wiewiora, E.},
  Booktitle                = {Proc. {Int. Conf. on Machine Learning} (ICML)},
  Year                     = {2009},
  Pages                    = {993--1000}
}

@inproceedings{tadic2004almost,
  title={Almost sure convergence of two time-scale stochastic approximation algorithms},
  author={Tadic, Vladislav B},
  booktitle={{IEEE} American Control Conf.},
  volume={4},
  pages={3802--3807},
  year={2004}
}

@article{taylor2009transfer,
  title={Transfer learning for reinforcement learning domains: A survey},
  author={Taylor, M. E. and Stone, P.},
  journal={Journal of Machine Learning Research},
  volume={10},
  number={Jul},
  pages={1633--1685},
  year={2009}
}

@article{teh2017distral,
  title={Distral: Robust Multitask Reinforcement Learning},
  author={Teh, Y. W. and Bapst, V. and Czarnecki, W. M. and Quan, J. and Kirkpatrick, J. and Hadsell, R. and Heess, N. and Pascanu, R.},
  journal={arXiv preprint arXiv:1707.04175},
  year={2017}
}

@misc{tieleman2012lecture,
  author        = {Tieleman, Tijmen and Hinton, Geoffrey},
  title         = {Lecture 6.5- {RMSP}rop: {D}ivide the gradient by a running average of its recent magnitude},
  year          = {2012},
  publisher={Coursera: Neural Networks for Machine Learning}
}

@inproceedings{tutunov2016exact,
  title={An exact distributed newton method for reinforcement learning},
  author={Tutunov, R. and Bou-Ammar, H. and Jadbabaie, A.},
  booktitle={IEEE Conf. on Decision and Control (CDC)},
  pages={1003--1008},
  year={2016}
}

@InCollection{van2012reinforcement,
  Title                    = {Reinforcement learning in continuous state and action spaces},
  Author                   = {Van Hasselt, Hado},
  Booktitle                = {Reinforcement Learning},
  Publisher                = {Springer},
  Year                     = {2012},
  Pages                    = {207--251}
}

@inproceedings{wei2012distributed,
  title={Distributed alternating direction method of multipliers},
  author={Wei, E. and Ozdaglar, A.},
  booktitle={IEEE Annual Conf. Decision and Control (CDC)},
  pages={5445--5450},
  year={2012}
}

@InProceedings{weinstein2012bandit,
  Title                    = {Bandit-Based Planning and Learning in Continuous-Action Markov Decision Processes.},
  Author                   = {Weinstein, Ari and Littman, Michael L},
  Booktitle                = {Int. Conf. on Automated Planning and Scheduling (ICAPS)},
  Year                     = {2012}
}

@Article{wierstra2014natural,
  Title                    = {Natural evolution strategies.},
  Author                   = {Wierstra, Daan and Schaul, Tom and Glasmachers, Tobias and Sun, Yi and Peters, Jan and Schmidhuber, J{\"u}rgen},
  Journal                  = {Journal of Machine Learning Research},
  Year                     = {2014},
  Number                   = {1},
  Pages                    = {949--980},
  Volume                   = {15}
}

@article{williams1992simple,
  title={Simple statistical gradient-following algorithms for connectionist reinforcement learning},
  author={Williams, R. J.},
  journal={Machine learning},
  volume={8},
  number={3-4},
  pages={229--256},
  year={1992},
  publisher={Springer}
}

@article{yaji2020stochastic,
  title={Stochastic Recursive Inclusions in Two Timescales with Nonadditive Iterate-Dependent Markov Noise},
  author={Yaji, Vinayaka G and Bhatnagar, Shalabh},
  journal={Mathematics of Operations Research},
  year={2020},
  publisher={INFORMS}
}

@inproceedings{parisotto2015actor,
  title={Actor-mimic: Deep multitask and transfer reinforcement learning},
  author={Parisotto, E. and Ba, J. L. and Salakhutdinov, R.},
  booktitle={Proc. Int. Conf. on Learning Representations (ICLR)},
  year={2016}
}

@InProceedings{zhang2018fully, 
    title = {Fully Decentralized Multi-Agent Reinforcement Learning with Networked Agents}, 
    author = {Zhang, Kaiqing and Yang, Zhuoran and Liu, Han and Zhang, Tong and Basar, Tamer}, 
    pages = {5872--5881}, 
    year = {2018}, 
    booktitle = {Proc. Int. Conf. on Machine Learning (ICML)},
}

@ARTICLE{zhao2015asynchronousI, 
author={X. Zhao and A. H. Sayed}, 
journal={{IEEE} Transactions on Signal Processing}, 
title={Asynchronous Adaptation and Learning Over Networks---Part I: Modeling and Stability Analysis}, 
year={2015}, 
volume={63}, 
number={4}, 
pages={811-826}, 
month={Feb}
}

@Article{zhao_performance_2012,
  Title                    = {Performance Limits for Distributed Estimation Over {LMS} Adaptive Networks},
  Author                   = {Zhao, X. and Sayed, A. H.},
  Journal                  = {{IEEE} Transactions on Signal Processing},
  Year                     = {2012},
  Number                   = {10},
  Pages                    = {5107--5124},
  Volume                   = {60},
}

@misc{PackerGao:1810.12282,
  Author = {Charles Packer and Katelyn Gao and Jernej Kos and Philipp Kr\"ahenb\"uhl and Vladlen Koltun and Dawn Song},
  Title = {Assessing Generalization in Deep Reinforcement Learning},
  Year = {2018},
  Eprint = {arXiv:1810.12282},
}


\begin{appendices}

\section{Numerical experiment details}
\label{app:numerical-experiments}

%
\subsection{Stability experiments (Sec. \ref{ssec:stability})}
%

\hspace{1em} \textbf{Inverted pendulum.}
The pendulum consists of a rigid pole and an actuated joint, with maximum torque clipped to the interval $[-2, 2]$. 
The pendulum starts at a random angle sampled from a uniform distribution on $[-\pi, \pi]$.
The goal is to take the pendulum to the upright position and balance there.
In the original single-task setting, pole mass and length are $(1.0, 1.0)$.
Our MRL problem consists of $25$ tasks with mass in $\{ 0.8, 0.9, 1.0, 1.1, 1.2 \}$,
and length in $\{ 0.8, 0.9, 1.0, 1.1, 1.2 \}$.

\textbf{Cart-pole swing-up.} 
We extend cart-pole balance to the case where the pole starts from the bottom and the task is to swing the pole to the upright position and balance. 
The reward function is $r = \frac{2}{1 + e^{d}} + \cos(\psi)$, 
where $d$ is the Euclidean distance of the pole from the upright position at the centre of the track, 
and $\psi$ is the pole angle.
The original single task uses parameters $(0.5, 0.25, 0.5)$ for the pole mass, pole half-length and cart mass respectively.
Our cart-pole swing-up MRL problem consists of $25$ tasks,
where pole mass is in $\{0.1, 0.2, 0.3, 0.4, 0.5 \}$,
pole half length is in $\{ 0.2, 0.4, 0.6, 0.8, 1.0 \}$,
and cart mass is $0.5$. 

We use $\gamma=0.99$ in all cases.

\textbf{Agent.} The actor and critic neural networks for both algorithms consist of $2$ hidden layers of $400$ neurons each with ReLU activation functions.
The output layer for the critic network is linear.
The output of the actor network includes a \textit{tanh} activation function that determines the mean of a normal distribution, and a \textit{Softplus} activation function that determines the variance for the normal distribution. 
We have also included a penalty in the loss function equal to the entropy of the policy, with penalty coefficient $0.0005$.

We use the ADAM optimiser \citep{kingma2015adam},
with learning rate $0.01$ for the critic and $0.001$ for the actor. 
Diff-SiAC performs a learning step ($i\leftarrow i+1$) every fifth episode.

%
\subsection{Comparison with Dist-MTLS (Sec. \ref{ssec:comparison-dist-mtls})}
%

\hspace{1em} \textbf{Cart-pole balance.} 
We use the OpenAI Gym \citep{OpenAI2016gym} implementation,
but with continuous force. The action follows a Gaussian distribution with mean in the interval $[-10, 10]$.
The episode finishes when either the pole is beyond $12$ degrees from vertical, 
the cart moves more than $2.4$ units from the centre,
or the simulation reaches $200$ time-steps.
The original single-task setting uses parameters $(0.1, 0.5, 1.0)$
for the pole mass, pole half-length and cart mass, respectively.
Our MRL problem consists of $25$ tasks with the following parameters:
pole mass in $\{0.1, 0.325, 0.55, 0.775, 1 \}$,
pole length in $\{ 0.05, 0.1625, 0.275, 0.3875, 0.5 \}$,
and cart mass $1$ in all cases.

We also use $\gamma=0.99$ in this problem.

\textbf{Agent.} We use the same agent architecture and hyperparameters as described in Appendix A.1. above for the stability experiments from Sec. \ref{ssec:stability}.

%
\subsection{Regularisation (Sec. \ref{ssec:regularisation})}
%

\hspace{1em}\textbf{Acrobot.} We use the Extreme Acrobot multitask environment introduced by \cite{PackerGao:1810.12282}. Based on the original OpenAI Gym implementation of the Acrobot environment \citep{OpenAI2016gym}, this setting is as the original Acrobot environment but the robot's mass, length and inertia are all sampled from independent uniform distributions on $[0.5, 0.75]\cup[1.25, 1.5]$. These environment parameters are sampled at the start of each experiment and fixed throughout the duration of training. In the original single-task setting, the robot's mass, length and inertia parameters are all set to 1.
The agent is penalised by receiving a reward of -1 for each time step until it reaches the goal state. 
For more details see \citep{PackerGao:1810.12282}.

\textbf{MuJoCo Hopper.} Inspired by \cite{PackerGao:1810.12282} we adapt the continuous control MuJoCo Hopper environment to develop a continuum of environments. We implement two separate extensions, one where the agent is in a setting with a randomly chosen constant wind and another where the agent's mass is scaled by a random value. In the setting with wind, we first adapt the medium through which the agent moves to have the same density and viscosity as air. Then, wind is implemented in the xy-plane. Wind is therefore a random 2-vector in which each element is sampled from an independent Beta distribution on the interval $[-15, 15]$. Once sampled, the direction and strength of the wind are held constant. The Beta distributions are parameterised with $\alpha=\beta=0.4$ in order to encourage more extreme conditions. We found that this was necessary to make the environment meaningfully different from the standard hopper environment.
In the environment where the agent's mass is randomly scaled, a scalar is sampled at the environment's inception and is used to scale the agent's mass. This scalar is fixed thereafter. The scalar is sampled from a Beta distribution transformed to be over the interval $[0.05, 0.95]$. We use $\alpha=\beta=0.1$ to again encourage more extreme settings and ensure a meaningfully different setting to that of the unmodified hopper environment.

We again use a discount factor, $\gamma = 0.99$ in these settings.

\textbf{Agent.} The actor and critic are both implemented as neural networks with 2 hidden layers of 64 units with \textit{tanh} activation function. The output layer is linear for both actor (producing logits parameterising the action distribution) and critic. 

To allow for fair comparison across the different agents, we tuned the hyperparameters of training for both Diff-A2C and Centralised A2C separately. The specialised agents use the hyperparameters for Centralised A2C as they are the same agents but without environment randomisation. Under Diff-A2C each agent only has one environment, as opposed to Centralised A2C where the agent samples experience from 25 environments in parallel. We found that Diff-A2C was most effective with a lower learning rate and a greater number of samples between updates when compared to Centralised A2C. This is consistent with findings of recent works on fully distributed RL \citep{assran2019gossip}. %
In all the experiments of Sec. \ref{ssec:regularisation}, the agents are trained with the RMSProp optimiser \citep{tieleman2012lecture} with $\alpha=0.99$ and $\epsilon=0.00001$. There is an entropy term included in the loss function with coefficient $0.01$.

Through tuning the learning rate for Centralised A2C and Diff-A2C independently we arrived at the following hyperparamers used in all experiments relating to Acrobot or MuJoCo Hopper.
Centralised A2C: Learning Rate 0.002, 5 steps per update. 
Diff-A2C: Learning Rate 0.0007, 60 steps per update.

For all the experiments in this section, the connectivity matrix $C$ used in Diff-A2C represents a ring, with every agent connected with two neighbours, such that equal weight is given to parameters from each agent involved in the combination step. 

\textbf{Generalisation.} The experiments used to consider generalisation in Acrobot found that the different sampled environments for Acrobot vary in difficulty as noted by the average performance of all agents during evaluation. In light of this, we chose an easy and a hard Acrobot environment to evaluate on. The easy environment has a robot arm length of 0.7046, an agent mass of 0.5259 and an inertia parameter of 0.6346. Contrastingly, the hard environment has a robot arm length of 1.3963, an agent mass of 1.3929 and an inertia parameter of 0.6256. We refer the interested reader to the work of \cite{PackerGao:1810.12282} for a detailed discussion regarding the influence of these environment parameters.
Table \ref{tab:acrobot-generalisation-relative} presents the results from Table \ref{tab:acrobot-generalisation} of section \ref{ssec:regularisation} but explicitly converted to performance relative to the specialised A2C agents (which represent 0\% in all cases).

\begin{table}
\centering
\begin{tabular}{lccc}
\toprule
                   & \wrap{Training\\Environment}     & \wrap{Out-of-Sample\\Easy Environment} & \wrap{Out-of-Sample\\Hard Environment} \\
\midrule
Diff-A2C           & \textbf{-33.2\%} (-55.9\%, -10.6\%) &  \textbf{31.3\%} (19.2\%, 43.3\%)   & \textbf{25.1\%} (12.8\%, 37.5\%)    \\
Centralised A2C    & -40.8\% (-68.7\%, -12.8\%) &  24.7\% (10.4\%, 39.0\%)   & 13.4\% (-1.5\%, 28.4\%)    \\
\bottomrule
\end{tabular}
\vspace*{0.2cm}
\caption{Zero-shot learning on Acrobot. Here we show performance relative to the Specialised Agents. 
Negative values show the performance gap, which is the case on the training environments.
Positive values denote outperforming the specialised agents, which is the case for out-of-sample environments. For the Specialised and Diff-A2C agents, a single agent from each experiment was chosen at random and evaluated on the environment they were trained on as well as the same held out easy and hard environments. Values in brackets denote a 95\% confidence interval. The values are averages over  8 seeds, 10 episodes each.}
\label{tab:acrobot-generalisation-relative}
\end{table}

%
\subsection{Network dependence (Sec. \ref{ssec:network-dependence})}
%

\hspace{1em} \textbf{Network topology.} 
We use the same implementation of the Cart-pole balance described in Appendix \ref{app:numerical-experiments}.2, but with the original single task parameters $(0.1, 0.5, 1.0)$ for the pole mass, pole half-length and cart mass, respectively.
We use the same agent described in Appendix \ref{app:numerical-experiments}.1.

\textbf{Noisy links.} 
We use the original (single-task) Gym implementation of Acrobot \citep{OpenAI2016gym}.
We use the same agent described in Appendix \ref{app:numerical-experiments}.3 again with agents arranged in a ring topology.

\section{Stability experiments in the single task problems}
\label{app:singletask}
\begin{figure}
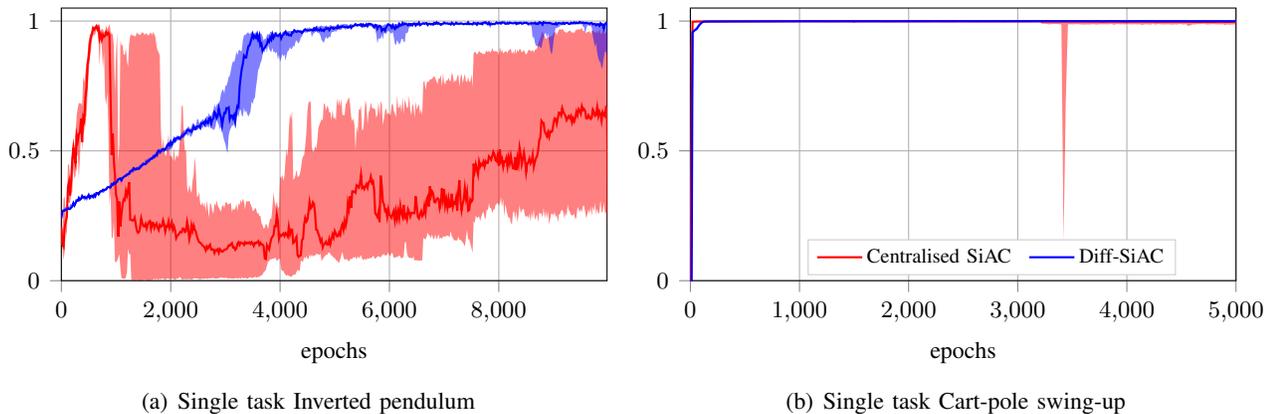

\subfigure[Single task Inverted pendulum]{\centering \input{pendulum_singletask_25_4_connectivity.tikz}}
\subfigure[Single task Cart-pole swing-up]{\centering \input{cartpoleswingup_singletask_25_4_connectivity.tikz}}
\caption{Stability in single task. Learning curves from training with 25 copies of the standard version of the Inverted pendulum (left) and Cart-pole swing-up (right) environments. Plots show median and first and third quartiles of the return averaged over all tasks, and 6 seeds.}
\label{fig:singletask_inverted_pendulum-swing_up}
\end{figure}

In this Appendix, we compare Centralised SiAC and Diff-SiAC in the standard single task problem for the Inverted pendulum and Cart-pole swing-up environments. Figure \ref{fig:singletask_inverted_pendulum-swing_up} shows that Centralised SiAC follows the same trend as for the multitask setting, it learns faster but is more unstable in both problems, especially in Inverted pendulum. 
On the other hand, Diff-SiAC learns robustly in both environments despite the highly correlated advantage estimates at each agent.
This result further supports our hypothesis that diffusion introduces a stabilising effect in the parameter update rule. 

\section{Proofs}
\label{app:proofs}

%

\subsection{Proof of Theorem \ref{theorem:policy-gradient}}

We expand the chain rule for vector functions, where $\left\langle\cdot, \cdot\right\rangle$ denotes inner product:
\begin{IEEEeqnarray}{rCl}
	\nabla_w \widetilde{L}	(v,w)
&=&
	\left\langle
		\nabla_w d_w
	,
		\frac{\partial L(v,w)}{\partial d_w}
	\right\rangle
\notag\\
&=&
	\int_{\St} \int_{\Ac}
		\nabla_w d(s,a)
		\frac{\partial L(v,w)}{\partial d_w(s,a)}
		\ ds\ da
\notag\\
&
\overset{(a)}{=}
&
	\int_{\St} \int_{\Ac}
			\left(
				\rho_\gamma^{\pi_w}(s)
				\nabla_{w} \pi_{w}(a|s)
				+
				\pi_{w}(a|s)
				\nabla_{w} \rho_\gamma^{\pi_w}(s)
			\right)
			\overline{A}(s,a)
		\ ds\ da
\notag\\
&
\overset{(b)}{=}
&
	\int_{\St} 
		\rho_\gamma^{\pi_w}(s)
		\int_{\Ac}
			\nabla_{w} \pi_{w}(a|s)
			\overline{A}(s,a)
	\ ds\ da
\notag\\
&
\overset{(c)}{=}
&
	\int_{\St} 
		\rho_\gamma^{\pi_w}(s)
		\int_{\Ac}
			\pi_{w}(a|s)
			\nabla_{w} \log \pi_{w}(a|s)
			\overline{A}(s,a)
	\ ds\ da
.
\end{IEEEeqnarray}
In step $(a)$, we expanded the gradient of the product.
In $(b)$, we used Bellman equation that says that the expected value of the advantage function over the policy distribution is zero, so that:
\begin{IEEEeqnarray}{rCl}
	\nabla_{w} \rho_\gamma^{\pi_w}(s)
	\int_\Ac
		\pi_{w}(a|s)
		\overline{A}(s,a)
		\ da
&=&
	0
.
\end{IEEEeqnarray}
In $(c)$, we used a standard reparameterisation trick:
$
	\nabla_{w} \pi_{w}(a|s)
=
	\pi_{w}(a|s) \nabla_{w} \log \pi_{w}(a|s)
$.
\hfill\IEEEQEDhere
%
%
%

\subsection{Proof of Theorem \ref{theorem:asymptotic-agreement}}

Let $M \defeq M_v + M_\pi$ be the length of each agent's parameter vector, $\bm{\phi}_{k,i}$,
and let $I_M$ denote the identity matrix of size $M$.
We introduce a network wide combination matrix:
$
\mc{C}
\defeq
    C \otimes I_{M}
$, where $\otimes$ denotes the Kronecker product.

In order to prove Theorem \ref{theorem:asymptotic-agreement}, we first present a useful lemma. For this, we introduce the following \textit{agreement} and \textit{disagreement} projection matrices, which will be used to compute the average and the difference from the average of a set of vectors (stacked on top of each other in a long vector), respectively:
\begin{IEEEeqnarray}{rCl}
	\Pagr
&
\defeq
&
    \frac{\1_N \1_N^\T}{N} \otimes I_M
,\quad
    \Pdis
\defeq
    I_{MN} - \Pagr
=
    I_{MN} - \frac{\1_N \1_N^\T}{N} \otimes I_M
,
\label{eq:projection-operators}
\end{IEEEeqnarray}
%
%
\begin{lemma}
Let $C$ be primitive and doubly stochastic. Then:
$\| \Pdis \mc{C} \| < 1$.
\label{lemma:bounded-deflated-C}
\end{lemma}
\begin{IEEEproof}
We can easily obtain the following equivalence:
\begin{IEEEeqnarray}{rCl}
    \Pdis
    \left(C \otimes I_M \right)
&=&
    \left(
        C - \frac{\1_N\1_N^\T}{N} 
    \right)
    \otimes I_M
.
\end{IEEEeqnarray}
Let $\lambda_n$, $u_n$, and $v_n$ denote the $n$-th eigenvalue, and $n$-th right and left eigenvectors of $C$, respectively.
From Assumption \ref{ass:connectivity-matrix} we have:
$
    C 
= 
    \frac{\1_N\1_N^\T}{N}
    +
    \sum_{n=2}^N
    \lambda_n u_n v_n^\T
$.
Hence, we bound the 1-norm:
\begin{IEEEeqnarray*}{rCl}
    \left\|
        C 
        - 
        \frac{\1_N\1_N^\T}{N}
    \right\|_1
&=&
    \left\|
        \sum_{n=2}^N
        \lambda_n u_n v_n^\T
    \right\|_1
<
    \left\|
        \frac{\1_N\1_N^\T}{N}
        +
        \sum_{n=2}^N
        \lambda_n u_n v_n^\T
    \right\|_1
=
    \left\|
        C 
    \right\|_1
=
    1
.
\end{IEEEeqnarray*}
We obtain the same result for infinity norm. We conclude by applying a well known inequality for the 2-norm \citep[Corollary 2.3.2]{golub1996matrix}:
$
    \left\|
        \Pdis \mc{C}
    \right\|
\le
\sqrt{
    \left\|
        \Pdis \mc{C}
    \right\|_1
    \left\|
        \Pdis \mc{C}
    \right\|_\infty
}
<
    1
$.
\end{IEEEproof}
%
%
%

We adapt previous works \citep[Ch. 4.6.2.]{bianchi_convergence_2013, valcarcel2017phdthesis} to our case.
Let's start introducing the following network-wide variables:
\begin{IEEEeqnarray}{rCl}
    \bm{\phi}_{i}
\defeq 
    \left(
    \begin{array}{c}
         \bm{\phi}_{1,i}\\
         \vdots \\
         \bm{\phi}_{N,i}
    \end{array}
    \right)
,\quad
    \mc{G}(\bm{\phi}_i)
\defeq
    \left(
    \begin{array}{c}
         g_1\left( \bm{\phi}_{1,i} \right)\\
         \vdots \\
         g_N\left( \bm{\phi}_{N,i} \right)
    \end{array}
    \right)
,\quad
    \bm{\mc{F}}_{i+1}
\defeq
    \left(
    \begin{array}{c}
         \bm{F}_{1+1,i}\\
         \vdots \\
         \bm{F}_{N+1,i}
    \end{array}
    \right)
.
\end{IEEEeqnarray}
Now, we can stack all agents' recursions as follows:
\begin{IEEEeqnarray}{rCl}
	\bm{\phi}_{i+1}
&
=
&
    \mc{C}
    \left(
    	\bm{\phi}_{i}
    	-
    	\alpha_{i+1}
    	\left(
    		\mc{G}\left(\bm{\phi}_{i}\right)
    		+
    		\bm{\mc{F}}_{i+1}
    	\right)
	\right)
.
\label{eq:network-recursion}
\end{IEEEeqnarray}
To aid our explanation, we introduce network-wide \textit{agreement} and \textit{disagreement} vectors:
\begin{IEEEeqnarray}{rCl}
    \bm{\phi}^{\rm agr}_i
&
\defeq
&
	\Pagr
	\bm{\phi}_i
=
(\1_N \otimes I_M) \overline{\bm{\phi}}_i
,\quad
    \bm{\phi}^{\rm dis}_i
\defeq
	\Pdis
	\bm{\phi}_i
=
    \bm{\phi}_i
    -
    (\1_N \otimes I_M) \overline{\bm{\phi}}_i
,
\label{eq:agreement-vectors}
\end{IEEEeqnarray}
Note that $\bm{\phi}_i$, $\bm{\phi}^{\rm agr}_i$ and $\bm{\phi}^{\rm dis}_i$ have length $MN$, 
while $\bm{\phi}_{k,i}$ and $\overline{\bm{\phi}}_i$ have length $M$.

From \eqref{eq:agreement-vectors}, we have:
$
    \bm{\phi}_i 
= 
    \bm{\phi}^{\rm agr}_i
    +
    \bm{\phi}^{\rm dis}_i
$.
Hence, our goal is to show: 
$\|\bm{\phi}^{\rm dis}_i\| \rightarrow 0$ $a.s.$, 
so that $\bm{\phi}_i \rightarrow \bm{\phi}^{\rm agr}_i$.
We build a recursion on $\bm{\phi}^{\rm dis}_i$ by multiplying both sides of \eqref{eq:network-recursion} with $\Pdis$:
\begin{IEEEeqnarray}{rCl}
	\bm{\phi}^{\rm dis}_{i+1}
&
=
&
    \Pdis
    \mc{C}
    \left(
    	\bm{\phi}^{\rm dis}_{i}
    	-
    	\alpha_{i+1}
    	\Pdis
    	\left(
    		\mc{G}\left(\bm{\phi}_{i}\right)
    		+
    		\bm{\mc{F}}_{i+1}
    	\right)
	\right)
,
\label{eq:disagreement-recursion}
\end{IEEEeqnarray}
where we used: $\Pdis \mc{C} = \Pdis \mc{C} \Pdis$.
Now, we bound $\|\bm{\phi}^{\rm dis}_{i+1}\|$:
\begin{IEEEeqnarray}{rCl}
	\left \| 
	    \bm{\phi}^{\rm dis}_{i+1} 
	\right \|
&
=
&
    \left \| 
        \Pdis
        \mc{C}
        \left(
        	\bm{\phi}^{\rm dis}_{i}
        	-
        	\alpha_{i+1}
        	\Pdis
        	\left(
        		\mc{G}\left(\bm{\phi}_{i}\right)
        		+
        		\bm{\mc{F}}_{i+1}
        	\right)
    	\right)
	\right \|
\notag\\
&
\le
&
    \left \| 
        \Pdis
        \mc{C}
	\right \|
	\left \| 
    	\bm{\phi}^{\rm dis}_{i}
    	-
    	\alpha_{i+1}
    	\Pdis
    	\left(
    		\mc{G}\left(\bm{\phi}_{i}\right)
    		+
    		\bm{\mc{F}}_{i+1}
    	\right)
	\right \|
\notag\\
&
<
&
    \left \| 
    	\bm{\phi}^{\rm dis}_{i}
    	-
    	\alpha_{i+1}
    	\Pdis
    	\left(
    		\mc{G}\left(\bm{\phi}_{i}\right)
    		+
    		\bm{\mc{F}}_{i+1}
    	\right)
	\right \|
\notag\\
&
\le
&
    \left \| 
    	\bm{\phi}^{\rm dis}_{i}
    \right \|
    +
    \alpha_{i+1}
    \left \| 
    	\Pdis
    	\left(
    		\mc{G}\left(\bm{\phi}_{i}\right)
    		+
    		\bm{\mc{F}}_{i+1}
    	\right)
	\right \|
,
\label{eq:bound-disagreement-recursion}
\end{IEEEeqnarray}
where the first inequality is due to the submultiplicative property of the norm; the second inequality is due to Lemma \ref{lemma:bounded-deflated-C}; and the third one is the Minkowski's inequality.
Rearranging terms and squaring and taking the limit on both sides we have:
\begin{IEEEeqnarray}{rCl}
\lim_{i\rightarrow \infty}
\left(
	\left \| 
	    \bm{\phi}^{\rm dis}_{i+1} 
	\right \|
	-
	\left \| 
    	\bm{\phi}^{\rm dis}_{i}
    \right \|
\right)^2
&
\le
&
\lim_{i\rightarrow \infty}
    \alpha_{i+1}^2
    \left \| 
    	\Pdis
    	\left(
    		\mc{G}\left(\bm{\phi}_{i}\right)
    		+
    		\bm{\mc{F}}_{i+1}
    	\right)
	\right \|^2
\notag\\
&
\le
&
\lim_{i\rightarrow \infty}
    \alpha_{i+1}^2
	\| \Pdis \|^2
    \left(
        \left \| 
    		\mc{G}
    		\left(
    		    \bm{\phi}_{i}
    		\right)
        \right\|^2
		+
        \left \| 
    		\bm{\mc{F}}_{i+1}
    	\right\|^2
    \right)
\notag\\
&=&
    0
\quad a.s.,
\label{eq:bound-squared-norm-disagreement}
\end{IEEEeqnarray}
where we used the submultiplicative property of the norm and Minkowski's inequality for the second inequality; the bounded norm of the projector;
Assumptions \ref{ass:lipschitz-gradients}, \ref{ass:square_integrable} and \ref{ass:bounded-iterates} to bound the norms of the gradient and the noise;
and Assumption \ref{ass:stepsizes} to   converge to zero and average the noise.

We conclude that the sequence
$
    \left\{ 
        \left \| 
	    \bm{\phi}^{\rm dis}_{i} 
	    \right \| 
    \right\}_{i=0}^\infty
$ converges $a.s.$
Let us introduce a shorthand for this limit:
$
    \left \| 
	    \bm{\phi}^{\rm dis}_{\infty} 
    \right \| 
\defeq
    \lim_{i\rightarrow\infty}
        \left \| 
	        \bm{\phi}^{\rm dis}_{i} 
        \right \| 
$.
Squaring the second line of \eqref{eq:bound-disagreement-recursion}, 
taking the limit, applying Minkowski's inequality, and rearranging terms yields:
\begin{IEEEeqnarray}{rCl}
    \left \| 
	    \bm{\phi}^{\rm dis}_{\infty} 
    \right \|^2
    \left(
        1
        -
        \left \| 
            \Pdis
            \mc{C}
	    \right \|^2
    \right)
&
\le
&
    \lim_{i\rightarrow \infty}
    \alpha_{i+1}
    \left \| 
        \Pdis
        \mc{C}
    \right \|^2
    \left \| 
    	\Pdis
    	\left(
    		\mc{G}\left(\bm{\phi}_{i}\right)
    		+
    		\bm{\mc{F}}_{i+1}
    	\right)
	\right \|^2
\quad a.s.
\end{IEEEeqnarray}
From Lemma \ref{lemma:bounded-deflated-C} and using the same arguments as in \eqref{eq:bound-squared-norm-disagreement}, we conclude:
$
    \left \| 
	    \bm{\phi}^{\rm dis}_{\infty} 
    \right \|
=
    0
\;\; a.s.
$
\hfill\IEEEQEDhere

%
\subsection{Proof of Theorem \ref{theorem:convergence-to-stationary}}
%
Let's build a recursion on $\bm{\phi}^{\rm agr}_i$ by multiplying \eqref{eq:network-recursion} by $\Pagr$:
\begin{IEEEeqnarray}{rCl}
	\bm{\phi}^{\rm agr}_{i+1}
&
=
&
	\bm{\phi}^{\rm agr}_{i}
	-
	\alpha_{i+1}
	\left(
	    \Pagr
		\mc{G}\left(\bm{\phi}_{i}\right)
		+
		\Pagr
		\bm{\mc{F}}_{i+1}
	\right)
,
\label{eq:agreement-recursion}
\end{IEEEeqnarray}
where we used $\Pagr \mc{C} = \Pagr$.
We will use the following notation for the average difference between the average of the gradient of each agent's parameter and the gradient of the average parameter:
\begin{IEEEeqnarray}{rCl}
    \bm{r}_i 
&
\defeq
&
    \Pagr 
    \left( 
        \mc{G}\left(\bm{\phi}_{i}\right)
        -
        \mc{G}\left(\bm{\phi}^{\rm agr}_{i}\right)
    \right)
.
\end{IEEEeqnarray}
Further, we define the average of the blocks inside $\bm{r}_i$, which are all equal:
\begin{IEEEeqnarray}{rCl}
    \overline{\bm{r}}_i 
\defeq
    \frac{1}{N}
    \1_N^\T \otimes I_M
    \bm{r}_i
.
\end{IEEEeqnarray}
Note that $\bm{r}_i$ and $\overline{\bm{r}}_i$ are random vectors of length $MN$ and $M$, respectively.
Finally, we introduce the average Martingale noise:
\begin{IEEEeqnarray}{rCl}
    \overline{\bm{F}}_{i+1} 
&
\defeq
&
    \frac{1}{N}
    \sum_{k=1}^N
        \bm{F}_{k,i+1}
.
\end{IEEEeqnarray}
Rearranging terms in \eqref{eq:agreement-recursion}, and from \eqref{eq:agreement-vectors}, we see that convergence of \eqref{eq:agreement-recursion} is equivalent to convergence of:
\begin{IEEEeqnarray}{rCl}
	\overline{\bm{\phi}}_{i+1}
&
=
&
	\overline{\bm{\phi}}_{i}
	-
	\alpha_{i+1}
	\left(
		\overline{g}\left(\overline{\bm{\phi}}_{i}\right)
		+
		\overline{\bm{r}}_i 
		+
		\overline{\bm{F}}_{i+1} 
	\right)
,
\label{eq:average-recursion}
\end{IEEEeqnarray}
where $\overline{g}$ is given by \eqref{eq:average-parameter-gradient}.
From Assumptions \ref{ass:lipschitz-gradients}, \ref{ass:stepsizes} and \ref{ass:bounded-iterates}, we conclude that:
$\lim_{i\rightarrow\infty} \alpha_{i+1} \overline{\bm{r}}_i = 0$ $a.s.$
Thus, the convergence of \eqref{eq:network-recursion} is equivalent to the convergence of this average recursion:
\begin{IEEEeqnarray}{rCl}
	\overline{\bm{\phi}}_{i+1}
&
=
&
	\overline{\bm{\phi}}_{i}
	-
	\alpha_{i+1}
	\left(
		\overline{g}\left(\overline{\bm{\phi}}_{i}\right)
		+
		\overline{\bm{F}}_{i+1} 
	\right)
.
\label{eq:average-recursion}
\end{IEEEeqnarray}

At this point, we can rely on Assumptions \ref{ass:lipschitz-gradients}--\ref{ass:ode-policy} to use the ODE method for the analysis of two time-scales stochastic approximations, in particular Theorem 1.1-Remark 2 from \cite{borkar1997stochastic} concludes the proof (see also \citep[Ch. 6]{borkar2008stochastic}).

\section{Diff-DAC Pseudocode}
\label{app:pseudocode}


\begin{algorithm}
\caption{Diff-DAC. This algorithm generalises the Diff-DAC architecture and therefore has Diff-SiAC and Diff-A2C as variants. The algorithm runs in parallel at every agent $k=1,\ldots,N$.}
\label{alg:diff-dac}
\textbf{Input:} Training termination condition (\texttt{terminate\_training}),
	stop sampling condition  (\texttt{stop\_sampling}),
	maximum number of steps per update $T$,
    learning rate sequences $\left\{\alpha_i, \beta_i\right\}_{i=1}^{\infty}$.\\
\begin{algorithmic}[1]
	\STATE Initialise critic, $v_{\xi_{k,0}}$, and actor, $\pi_{w_{k,0}}$, networks, $\forall k \in \N$.
    \STATE Initialise iteration counter, $i = 0$.
    \STATE \textbf{while} not \texttt{terminate\_training}\textbf{:}
    \STATE \hspace{1em} Initialise empty trajectory, $\M_k = \{\}$.
    \STATE \hspace{1em} Initialise step counter: $t = 0$.
    \STATE \hspace{1em} \textbf{while} $t < T$ and not \texttt{stop\_sampling}\textbf{:}
    \STATE \hspace{2em} Observe environment state, $s_{k,t}$.
	\STATE \hspace{2em} Select action $a_{k,t} \sim \pi_{w_{k,t}}(\cdot | s_{k,t})$.
    \STATE \hspace{2em} Execute $a_{k,t}$ and observe $r_{k,t+1}$ and $s_{k,t+1}$.
    \STATE \hspace{2em} Store tuple $(s_{k,t}, a_{k,t}, r_{k,t+1}, s_{k,t+1})$ in $\M_k$.
	\STATE \hspace{2em} Update step counter: $t \leftarrow t + 1$.
	\STATE \hspace{1em} \textbf{end while}
    \STATE \hspace{1em} \textbf{for} each sample $t \in \M_{k}$\textbf{:}
    \STATE \hspace{2em} Compute advantage function (see Equations \eqref{eq:advantage-estimator} and \eqref{eq:a2c-advantage})
	\STATE \hspace{1em} \textbf{end for}
    \STATE \hspace{1em} Compute distributed critic update:
		\begin{IEEEeqnarray*}{rCl}
			\widehat{\xi}_{k,i+1}
		&
		=
		&
		    \texttt{SGD\_optimizer}
		    \left(
		        \alpha_{i+1}
		        ,\:
		        \xi_{k,i}
		        ,\:
		        \frac{1}{|\M_{k}| - 1}
                \sum_{t=0}^{|\M_{k}|}
				    \nabla_\xi v_{\xi_{k,i}} \left( s_{k,t} \right)
				    \widehat{A}_{k,t}
		    \right)
\qquad\qquad\qquad
		\\
			\xi_{k,i+1}
		&
		=
		&
			\sum_{l \in \N_k}
				c_{lk}
				\widehat{\xi}_{l,i+1}
		\end{IEEEeqnarray*}

    \STATE \hspace{1em} Compute distributed actor update:
		\begin{IEEEeqnarray*}{rCl}
			\widehat{w}_{k,i+1}
		&
		=
		&
		    \texttt{SGD\_optimizer}
		    \left(
		        \beta_{i+1}
		        ,\:
		        w_{k,i}
		        ,\:
		        \frac{1}{|\M_{k}|}
                \sum_{t=0}^{|\M_{k}| - 1}
				    \nabla_{w} \log \pi_{w_{k,i}}\left(a_{k,t}|s_{k,t}\right)
				    \widehat{A}_{k,t}
		    \right)
\qquad
		\\
			w_{k,i+1}
		&
		=
		&
			\sum_{l \in \N_k}
				c_{lk}
				\widehat{w}_{l,i+1}
		\end{IEEEeqnarray*}
	\STATE \hspace{1em} Update iteration counter: $i \leftarrow i +1$.
    \STATE \textbf{end while}
\end{algorithmic}
\textbf{Return:} Critic and actor weights: $\xi_{k,{\rm end}}$, $w_{k,{\rm end}}$.
\end{algorithm}

Algorithm \ref{alg:diff-dac} shows the pseudocode for the Diff-DAC architecture, which can be adapted to multiple actor-critic algorithms, including Diff-SiAC and Diff-A2C among others.
The choice of advantage function, the termination conditions (for training and sampling) and the optimiser differentiates our implementations of Diff-SiAC and Diff-A2C. 

Under Diff-SiAC the advantage function in \eqref{eq:advantage-estimator} is used,
while Diff-A2C uses the more advanced advantage estimator in \eqref{eq:a2c-advantage}.

Diff-SiAC also terminates training after a fixed number of episodes and updates the parameters at the end of each episode. In this case, the termination condition is based on an episode count. Our implementation of Diff-A2C however, terminates training and sampling based on the number of steps taken in the environment. 

At each iteration, Diff-SiAC resets the environment, takes samples until the end of the episode or when reaching a maximum number of environment steps, and then updates parameters.
On the other hand, Diff-A2C updates at a fixed interval in terms of environment steps and may, therefore, encounter terminal states during data collection. In such cases, data collection continues while marking the terminal states in the trajectory $\M_k$ to aid in calculating the advantage estimates. We note that, in the case of an environment with fixed episode length, it is trivial to fix the termination and update interval conditions to be equivalent for Diff-SiAC and Diff-A2C.

\texttt{SGD\_optimizer} is a placeholder for different stochastic gradient descent variants, that take the learning rate, current parameter value, and gradient estimate and output the updated parameter. 
Diff-SiAC uses Adam \citep{kingma2015adam}, while Diff-A2C uses RMSProp \citep{tieleman2012lecture}. We note that the results do not change significantly when using Adam for Diff-A2C.

The implementation of Diff-SiAC uses a single process, so all agents update synchronously.
The implementation of Diff-A2C however exploits multiprocessing. We have performed all simulations on a single machine with a single GPU. Due to the multiprocessing, each Diff-A2C agent performs its local update asynchronously.
Hence, some of its neighbours might not be ready when an agent begins the diffusion update.
In this scenario, we allow each agent to perform its diffusion update with the most recently received values for the neighbours' parameters, provided that no value for a neighbour's parameters is more than 5 iterations old. Where this staleness limit is reached, an agent waits for a set of up-to-date values from its lagging neighbour(s). 
In practice, this limit of 5 update iterations of lag between neighbours is very rarely hit and our results are therefore not sensitive to this value. For the experiments exploring the sensitivity of Diff-A2C to noisy links in Sec. \ref{ssec:network-dependence}, we set this limit to zero so agents do not wait for their neighbours. Our implementation of Diff-A2C adapts the code put forward by \cite{assran2019gossip}.

\end{appendices}

\end{document}